\documentclass[times]{elsarticle}
\usepackage{lmodern}
\usepackage{moreverb} 
\usepackage[colorlinks,bookmarksopen,bookmarksnumbered,citecolor=red,urlcolor=red]{hyperref}
\usepackage{amsfonts,amsmath,amssymb}
\usepackage{psfrag,pstricks,pst-node}
\usepackage{mathrsfs}
\usepackage{psfrag}  
\usepackage[ruled,vlined]{algorithm2e}
\usepackage{fancyhdr}  
\usepackage{psfrag} 
\usepackage{commath}
\usepackage{enumerate} 
\usepackage{algorithmic}
\usepackage{bm}
\usepackage{booktabs}
\usepackage[margin=2.5cm]{geometry}
\newtheorem{remark}{Remark}

\newcommand\BibTeX{{\rmfamily B\kern-.05em \textsc{i\kern-.025em b}\kern-.08em
		T\kern-.1667em\lower.7ex\hbox{E}\kern-.125emX}}

\begin{document}
	
	\begin{frontmatter}
		\renewcommand{\thefootnote}{\fnsymbol{footnotemark}}
		
		\fancypagestyle{plain}{%
			\fancyhf{} 
			\fancyhead[RO,RE]{\thepage} 
		}
		
		\title{Parametric Taylor series based latent dynamics identification neural networks}
            \author[lab1]{Xinlei Lin}
		\author[lab1]{Dunhui Xiao\corref{cor1}}
		\cortext[cor1]{Corresponding author}
		\ead{xiaodunhui@tongji.edu.cn}   
		\address[lab1]{School of Mathematical Sciences, Key Laboratory of Intelligent Computing and Applications(Ministry of Education), Tongji University, Shanghai 200092, CHINA}

\begin{abstract}
Numerical solving parameterised partial differential equations (P-PDEs) is highly practical yet computationally expensive, driving the development of reduced-order models (ROMs). 
Recently, methods that combine latent space identification techniques with deep learning algorithms (e.g., autoencoders) have shown great potential in describing the dynamical system in the lower dimensional latent space, for example, LaSDI, gLaSDI and GPLaSDI. 

In this paper, a new parametric latent identification of nonlinear dynamics neural networks, P-TLDINets, is introduced, which relies on a novel neural network structure based on Taylor series expansion and ResNets to learn the ODEs that govern the reduced space dynamics. 
During the training process, Taylor series-based Latent Dynamic Neural Networks (TLDNets) and identified equations are trained simultaneously to generate a smoother latent space. 
In order to facilitate the parameterised study, a $k$-nearest neighbours (KNN) method based on an inverse distance weighting (IDW) interpolation scheme is introduced to predict the identified ODE coefficients using local information. 
Compared to other latent dynamics identification methods based on autoencoders, P-TLDINets remain the interpretability of the model. Additionally, it circumvents the building of explicit autoencoders, avoids dependency on specific grids, and features a more lightweight structure, which is easy to train with high generalisation capability and accuracy. Also, it is capable of using different scales of meshes.  
P-TLDINets improve training speeds nearly hundred times
compared to GPLaSDI and gLaSDI, 
maintaining an $L_2$ error below $2\%$ compared to high-fidelity models.
\end{abstract}
		
	\end{frontmatter}
	
\section{Introduction}
\label{Section1}
\vspace{-2pt}


In the fields of engineering and physics, the behaviours of many complex systems are governed by partial differential equations (PDEs)\cite{negri2015efficient, benner2015survey,  wang2018model}. To describe and understand these systems through PDEs, physical simulations and computational fluid dynamics (CFD) provide essential guidance for advanced design and practical industry applications\cite{roy2005review, katsikadelis2014numerical}. 
During the past few decades, numerical methods such as the finite element method (FEM)\cite{gallagher1975finite} and the finite volume method (FVM)\cite{eymard2000finite} have been continuously advancing, while significant progress has been made in computer hardware and GPUs simultaneously. However, as the demand for superior accuracy in discrete solutions increases and grids become highly refined, the computational requirements of these traditional methods become exceedingly considerable. This is especially evident for problems involving multiple repeated simulations, such as parametric problems\cite{xiao2014non, pawar2020data}, where the expansion of modelling iterations leads to dramatic resource consumption. Consequently,  these issues may lead to limitations in effective application in practice. 

To overcome these limitations, reduced-order modelling (ROM) emerges as a promising approach that conquers the computational cost impediments in many areas, such as turbulent flows\cite{xiao2016non, fukami2021machine, fu2023non}, aerodynamics\cite{fukami2023grasping} digital twins\cite{wagg2020digital}, biomedical sciences\cite{cimrak2012modelling}, finance\cite{judd2014smolyak}, meteorology science\cite{li2009bibliometric}, geosciences\cite{karpatne2018machine}, aerospace\cite{eren2017model} and plasma dynamics\cite{nayak2021detection}. The primary objective of ROM is to reduce the dimensions of full-order models by projecting the system onto a lower-dimensional latent linear or non-linear subspace. Typically, ROM achieves remarkable speed enhancements compared to high-fidelity models, albeit with a negligible decrease in accuracy, rendering it incredibly attractive.
The parameterised nonlinear reduced order methods (P-NIROM)\cite{xiao2017parameterized} is a method designed for quick prediction and reliable simulation of parameterised partial differential equations (P-PDEs), e.g. the Navier-Stokes equations.  

Popular linear reduced-order methods include proper orthogonal decomposition (POD)\cite{berkooz1993proper}, Galerkin projection\cite{fang2013non}, dynamic mode decomposition (DMD)\cite{proctor2016dynamic} and reduced basis method (RB)\cite{maday2002reduced}. Autoencoders, a nonlinear projection technique in reduced-order models, demonstrate impressive capabilities in dimensionality reduction\cite{fu2023non, pichi2024graph}. 
They have already been widely applied in various fields such as aerodynamic wing design\cite{yonekura2021data}, fluid dynamics\cite{lee2020model}, Magneto hydrodynamics (MHD)\cite{kaptanoglu2021physics}, high-energy physics\cite{odyurt2024reduced} and long-span bridge building\cite{cui2023wind}. The combination of autoencoders with neural network architectures, such as convolutional neural networks (CNNs)\cite{fukami2020convolutional}, recurrent neural networks (RNNs)\cite{orvieto2023resurrecting}, long short-term memory networks (LSTMs)\cite{zhang2020physics},  residual neural networks (ResNets)\cite{he2016deep}, transformers\cite{solera2024beta},  DeepONets\cite{lu2021learning} etc., has shown high efficiency in reduced modelling and prediction tasks. 
In addition to autoencoders, there are other network structures that can achieve dimension reduction, such as generative adversarial networks (GANs)\cite{goodfellow2014generative}, t-distributed stochastic neighbour embedding (t-SNE)\cite{das2023t-sne}, transfer learning\cite{pan2008transfer} and Latent Dynamics Networks\cite{regazzoni2024ldnets}. 
In particular, Latent dynamic networks (LDNets)\cite{regazzoni2024ldnets} provide a brand new perspective in ROM community, which automatically learns and discovers low-dimensional features of dynamic systems, i.e., latent state variables.

The discovery of governing equations in latent spaces is a particularly promising approach to address the challenges since the dynamics in the latent space can be identified as a set of ordinary differential equations (ODEs). 
These identified governing equations can be easily solved and mapped back to the high-dimensional flow field solution space. 
In other words, these ODEs are better for revealing the underlying physical or mathematical laws of model learning, leading to improved model interpretability and enhanced generalisation. 
The sparse identification of nonlinear dynamics (SINDy)\cite{2016SINDy}, as a method of dynamics identification, through the use of a function library and sparse linear regression, has seen widespread applications and led to the development of several related parametric PDE algorithms\cite{champion2019data, he2023glasdi, fries2022lasdi, bonneville2024gplasdi}, such as PDE-FIND\cite{rudy2017data}, RE-FIND\cite{lin2024data}, LaSDI\cite{fries2022lasdi}, gLaSDI\cite{he2023glasdi} and GPLaSDI\cite{bonneville2024gplasdi}. 
By extracting coefficients from latent governing equations and combining them with interpolation or fitting methods, such as radial basis function (RBF) interpolation\cite{qiao2012identification}, Gaussian process regression (GPR)\cite{bonneville2024gplasdi}, or linear interpolation techniques, efficient and effective evaluation and prediction are easily achieved across the whole parameter space.


In this paper, inspired by the latent dynamics networks (LDNets) proposed by \cite{regazzoni2024ldnets}, a new Multiscale Parametric Taylor series-based Latent Dynamics Identification Neural Networks (P-TLDINets) are presented.
The P-TLDINets are constructed via Taylor series expansions, ResNets, and $k$-nearest neighbours (KNN) algorithm combined with an inverse distance weighting (IDW) interpolation schema. 
Apart from identifying latent equations like LDNets, P-TLDINets can be applied to more complex parametric PDE problems, for example, problems in which the initial conditions vary with the parameters. 
The LDNets can only solve parametric problems where the initial conditions do not change with the parameters. 
In addition, the P-TLDINets model uses only a small amount of high-fidelity training data to achieve accurate, robust, and efficient data-driven reduced-order modelling for physical systems. Furthermore, the P-TLDINets is a multiscale model which is capable of running simulations with variable meshes. This allows P-TLDINets to predict high-dimensional approximation data at any query point $\bm{x}$ in the computational domain $\Omega$, even if some of them may not be part of the training grids. This is another novelty for the newly presented P-TLDINets. 

The subsequent part of the paper is organised as follows:
Section \ref{sec:governing-equation} describes the related governing equations for parametric systems. The framework of P-TLDINets includes several parts, such as a family of networks called Taylor series-based Latent Dynamics Networks (TLDNets) in Section \ref{sec:TLDNets} 
and Identification of Dynamics (ID models) in latent space in Section \ref{sec:sindy}. 
Section \ref{sec:knn} shows the KNN method and the interpolating process. Section \ref{sec:data-preparation} is data preprocessing, including data sampling and normalisation of P-TLDINets. 
The whole framework of P-TLDINets is summarised in Section \ref{sec:framework}. 
The capabilities of P-TLDINets are demonstrated using two numerical examples: 2D-Burgers equations in \ref{sec:2d-burgers} and lock exchange in \ref{sec:lock-exchange}. Furthermore, we test the effectiveness of the P-TLDINets model on multiscale problems to show its greater versatility. 
Through these examples, P-TLDINets demonstrate their distinct differences and superiority over autoencoder-based latent dynamics identification methods. Finally, we summarise the main innovations and applications prospects of the proposed model, providing a comprehensive overview of this research in Section \ref{sec:conclusion}.

\section{Governing equations of parameterised dynamic systems}\label{sec:governing-equation}
\vspace{-2pt}

In practical simulations, the PDEs that govern the systems are typically characterised by the parameter vector $\bm{\mu}$, which contains initial or boundary conditions as well as the dynamics of the system. 
Suppose $\bm{\mu}$ is the input variable in a parameter domain $\mathcal{D} \subset \mathbb{R}^{N_{\mathcal{D}}}$, in which $N_{\mathcal{D}}$ is the number of parameters in a $\bm{\mu}$ and $N_{\mathcal{D}}$ can be arbitrary in realistic problems. In this work, a parameter space is considered to be $2D$. 
We define the governing PDEs of a parameterised dynamical system as 

\begin{equation}\label{equ:govern_equ}
\left\{
\begin{aligned}
    \bm{F}(\bm{u}(t, \bm{x}; \bm{\mu}), t, \bm{x}; \bm{\mu}) & = s(t, \bm{x}, \bm{\mu})  & \quad (t, \bm{x}) \in [0, T] \times \Omega \\
    \bm{u}(t=0, \bm{x} ;\bm{\mu}) & = \bm{u}_0(\bm{\mu}),  & 
\end{aligned}
\right.
\end{equation}
where $\bm{u}: [0, T] \times \Omega \mapsto \mathcal{U} \subset \mathbb{R}^{N_{f}}$ stands for either a scalar field or vector fields, such as velocity, pressure, or temperature. $T \in \mathbb{R}^+$ is the simulation time period and $\Omega \in \mathbb{R}^d$ defines the space domain of the problem. $\bm{F}:\mathbb{R}^{N_{f}} \times [0,T] \times \Omega \times \mathcal{D} \mapsto \mathbb{R}^{N_f}$ denotes the dynamic evolution of $\bm{u}$ and conversation laws, while $s$ is the source term.  $\bm{u}_0$ denotes the initial value of $\bm{u}$. 

In this paper, discrete points in space are denoted by $\bm{x} \in \Omega$, which are typically vectors of length determined by the problem's dimension $d$. For example, in a problem based on a two-dimensional space, $\bm{x} = (x_1, x_2)$. 
Usually, high-dimensional spatial and temporal discretisations of the problem in Equation \ref{equ:govern_equ} can be carried out by classical numerical methods, such as FEM and FVM. However, as the dimension of the solution $N_{\bm{u}}$ increases or the spatial domain $\Omega$ is characterised by a complicated geometry, solving Equation \ref{equ:govern_equ}  may require significant computational resources.  
Assume that the discrete solutions corresponding to a given parameter vector $\bm{\mu}_i\in \mathcal{D}$ are defined as $\bm{u}_i=[u_1, u_2, \dots, u_{N_{u}}] \in \mathbb{R}^{N_{u}}$. 

In this work, the P-TLDINets includes two components: Taylor series-based Latent Dynamic Networks (TLDNets) and Identification of Dynamics models (ID models). They are introduced in Section \ref{sec:TLDNets} and Section \ref{sec:sindy} respectively.

\begin{table}[!h]
\centering
\caption{Notations}
\begin{tabular}{cl}
   \toprule
   Notation & Description \\
   \midrule
   $\bm{u}$  &  The state variables of dynamics. \\
   $\mathcal{S} $ & The latent state space. \\
   $\Omega$ & The whole high-dimensional computational domain. And the coordinates $\bm{x} \in \Omega$. \\
   $N_{u}$& The number of spatial discretisation on $\Omega$. \\
   $N_s$ & The number of state variables, i.e. the dimension of latent space. \\
   $N_{\mathcal{D}}$ & The dimension of the entire parameter space $\mathcal{D}$. \\
   $N_{\bm{\mu}}$ & The number of discrete points in the training parameter space $\mathcal{D}_{train}$. \\
   $N_{h}$ & The number of discrete points in the testing parameter space $\mathcal{D}^h$. \\
   $N_t$ & The number of the discrete time domain.\\
   \bottomrule
\end{tabular}
\end{table}

\section{Taylor series-based Latent Dynamic Neural Networks (TLDNets)}\label{sec:TLDNets}

The Taylor series-based Latent Dynamic Networks (TLDNets) consist of three network structures, involving the dynamic network $NN_{dyn}$, the reconstruction network $NN_{rec}$, and the initial mapping network $NN_{z_0}$, corresponding to the model parameters $W_{dyn}$,  $W_{rec}$ and $W_{z_0}$ respectively. 
They are constructed using fully connected neural networks (FCNNs) and residual neural networks (introduced in Section \ref{sec:resnet}).
These three networks are used to compute the time derivatives of the latent state variables, reconstruct the scalar data $\bm{u}$, and generate the initial latent states respectively. 
The traditional latent dynamics networks (LDNets) consist of two neural networks for dynamics learning and reconstruction respectively, but cannot be used on parametric problems with initial values varying with parameters, which limits their capacity of generalisation. 
To solve this, a brand new FCNN structure $NN_{z_0}$, introduced in Section \ref{sec:ini} in detail, is added to TLDNets so that they are able to be adapt to more situations. Besides, ResNets and Taylor series expansion theory are integrated into TLDNets to enhance the representation ability of P-TLDINets, enabling their application to more complex fluid cases.

\begin{remark}\label{rmk:well-defined}
    Assume that the mapping from $\bm{\mu}$ to $\bm{u}$ is well-defined and $\bm{u}(x,t)$ has an underlying low-dimensional state $\bm{z}(t)=[z_1, z_2, \dots, z_{N_s}] \in \mathbb{R}^{N_s}$. 
    If $\bm{z}(t)$ is eventually restored by $NN_{rec}$ to obtain an estimated value $\tilde{\bm{u}}$ close to $\bm{u}$ enough, then this hypothesis can be deemed correct. 
\end{remark}

\subsection{The dynamic model and the reconstruction model}\label{sec:dyn and rec}

The latent space $\mathcal{S}$ is identified by the dynamic model and the initial mapping model. After obtained $\bm{z}(t)$ and $\dot{\bm{z}}(t)$ through $NN_{dyn}$ and $NN_{z_0}$,  ID models in Section \ref{sec:sindy} are then utilised to discover the corresponding governing equations. Finally the reconstructed high-dimensional flow field $\tilde{\bm{u}}$ is computed based on $NN_{rec}$ and latent ODE solutions of ID models. The specific whole formulas are as follows: 

\begin{equation}\label{equ:NN}
\left\{
\begin{aligned}
    \relax[\frac{d}{dt}\bm{z}(t), \frac{d^2}{dt^2}\bm{z}(t)] & = NN_{dyn}(\bm{z}(t), 
    \bm{\mu})  &   t \in(0, T]  \\
    \bm{z}(0) & = NN_{z_0}(\bm{\mu})  & \\
    \tilde{\bm{u}}(x,t) & = NN_{rec}(\bm{z}(t), \bm{\mu}, \bm{x}).  & t \in [0, T]
\end{aligned}
\right.
\end{equation}

Here, the role of $NN_{z_0}: \mathcal{D} \rightarrow \mathcal{S}$ is to map the parameter vector $\bm{\mu}$ to a initial low-dimensional state $\bm{z}(0)$. 
To better understand the contributions of each component within TLDNets, $NN_{dyn}$ and $NN_{rec}$  can be analogised to the generator and discriminator in generative adversarial networks (GANs)\cite{goodfellow2014generative}. It is assumed that $NN_{dyn}$ is capable of establishing a mapping between the high-dimensional snapshot data $\bm{u}$ and the hypothetical space $\mathcal{S}\subset \mathbb{R}^{N_s} (N_s \ll N_u)$. In other words, $NN_{dyn}$ computes the first-order time derivative $\frac{d}{dt}\bm{z}(t_{m+1})$ and the second-order time derivative $\frac{d^2}{dt^2}\bm{z}(t_{m+1})$ at the next time point based on the current state $\bm{z}(t_m)$ and the input $\bm{\mu}$. 
Subsequently, the low-dimensional state variable $\bm{z}(t) \in \mathcal{S}$ is calculated through a Taylor series summation and an iterative structure. The specific mathematical process is determined by: 

\begin{equation}\label{equ:dynamic_state}
\left\{
\begin{aligned}
    \bm{z}(t_0) & = NN_{z_0}(\bm{\mu}) \\
    \bm{z}(t_{m+1}) & = \bm{z}(t_m) +  \frac{d}{dt}\bm{z}(t_m) \Delta t +  \frac{d^2}{dt^2}\bm{z}(t_m) (\Delta t)^2 + o((\Delta t)^2). \quad m=1,\dots,N_t
\end{aligned}
\right.
\end{equation}
Once the differences between $\tilde{\bm{u}}$ and $\bm{u}$ in Equation \ref{equ:NN} is negligible, Remark \ref{rmk:well-defined} is considered to hold. 
Unlike dimensional reduction methods such as POD, the latent variables $\bm{z}(t)$ and the related space $\mathcal{S}$ are not predefined but are discovered during the training process.

In particular, in contrast to traditional dimension reduction methods, the structure of TLDNets ensures that its network parameters are significantly fewer compared to an autoencoder, leading to less training time and random access memory (RAM) utilisation rate. After training, P-TLDINets allow individual queries of $\bm{u}$ based on distinct coordinate information and the parameter vector $\bm{\mu}$, which means that it allows predictions at additional points $\tilde{\bm{x}}$ other than just at the training coordinates $\bm{x}$. Furthermore, a P-TLDINets model supports flexible coordinate input, thus, it can be classified as a mesh-free method. 

In TLDNets, a hyperbolic tangent (Tanh) nonlinear activation function is used in $NN_{dyn}$ and $NN_{rec}$. The network parameters in $NN_{dyn}$ are updated by backpropagation based on the reconstruction results from $NN_{rec}$.
The loss function of $NN_{rec}$ model is summarised as
\begin{equation}
    \begin{aligned}
       &  Loss_{rec}(W_{rec}, W_{dyn}) =   \\
        & \frac{1}{N_{\bm{\mu}}} \sum_{i=1}^{N_{\bm{\mu}}} \big[\frac{1}{N_t+1}\sum_{l=0}^{N_t}(\frac{1}{N_u}\sum_{m=1}^{N_u}Loss_{function}(NN_{rec}(\bm{z}(t_l), \bm{\mu}^i, x^m), u_m(t_l))) \big].
    \end{aligned}
\end{equation}

\subsection{The initial mapping model}\label{sec:ini}

The original network family LDNets in\cite{regazzoni2024ldnets} could only be applied to parameterised problems with the same initial condition. The fundamental reason lies in an assumption made therein: without loss of generality, the potential initial condition $\bm{z}(0) = 0$ is postulated in LDNets, based on the hidden properties of  $\bm{z}(t)$\cite{regazzoni2019s0}. 
The integration of $NN_{z_0}$ is employed to predict the initial state variable $\bm{z}(0)$ and promises that TLDNets will be invoked to more complex parameterised problems, especially on those with different initial conditions, thus expanding its applicability. 
The loss function of $NN_{z_0}$ is on the basis of mean square error (MSE) and is summarised as
\begin{equation}
    Loss_{z_0}(W_{z_0}) = MSE(NN_{rec}(\bm{z}_0, \bm{\mu}, \bm{x}), \bm{u}(t_0)).
\end{equation}
Since $NN_{z_0}$ is jointly trained with $NN_{dyn}$ and $NN_{rec}$, this is reflected in the total loss function (as discussed in Section \ref{sec:framework}). To achieve better training results, a weighting hyperparameter $\omega_{z_0}$ is imposed on $Loss_{z_0}$.


\subsection{Residual Learning neural Network (ResNet)}\label{sec:resnet}

Pure FCNNs have shown a considerable degree of superiority in function approximation across various fields\cite{devore2021neural}. However, as the depth of networks increases with the problem complexity of the architecture, deep learning models are prone to suffer from gradient vanishing and overfitting, resulting in weakened performance. The deep residual learning neural network (ResNet) is an architecture proposed to address these challenges to a certain extent, improving the ability of networks to capture more dynamics in data.

\begin{figure}[htbp]
    \centering
	\includegraphics[width=0.5\linewidth]{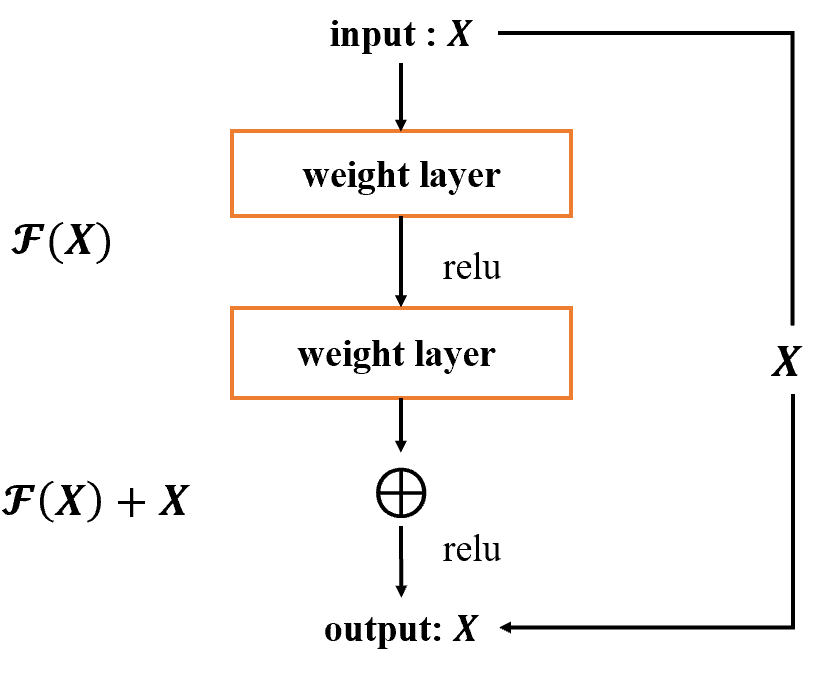}
    \caption{Residual learning neural network block structure}
    \label{fig:resnet}
\end{figure}

Instead of purely FCNNs, we use ResNet in the $NN_{dyn}$ and $NN_{rec}$ networks. ResNet was initially based on convolutional neural network (CNN) technology in the field of computer vision and image classification\cite{he2016deep}. Here, the residual learning concept is applied to FCNNs. 
The main difference between residual learning and classical FCNNs lies in the structure of their networks, as shown in Figure \ref{fig:resnet}. A FCNN consists of multiple hidden layers, where adjacent neurons are directly connected by weight matrices and biases, and then nonlinearity is provided by activation functions. In contrast, for a ResNet, there exists a shortcut path, connecting the input and output layers directly, known as a "skip connection." 

Assume that the input to a ResNet block is $X$, and set the corresponding mapping as $\mathcal{H}$. In Figure \ref{fig:resnet}, the weight layers consist of several linear neurons, and the activation function is the rectified linear unit (ReLU) function. Denote the mapping formed only by stacked nonlinear layers as $\mathcal{F}$. Then, the output of a single ResNet block can be expressed as 
\begin{equation}
    \mathcal{H}(X):=\mathcal{F}(X)+X.
\end{equation}

\section{Identification of dynamics (ID model) in the latent space}\label{sec:sindy}

Afterwards, ID models are applied in $\bm{z} \in \mathcal{S}$ to discover intrinsic characteristics of latent dynamics. Finally, according to the input spatial coordinate information $\bm{x} \in \Omega$, the reconstruction network $NN_{rec}$ maps to the original flow fields, receiving the predicted value $\tilde{\bm{u}}$. 
The vector $\bm{z}$  is equivalent to a compressed representation of the high-dimensional physical data $\bm{u}$, defined over both time and space. The temporally-dependent $\bm{z}$ reflects the physical features of the original solution manifold so that it can be viewed as a dynamical system governed by a set of ODEs.
Here, consider the reduced-order dynamic identification model as follows:
\begin{equation}\label{equ:sindy-govern}
    \frac{d}{dt} \bm{z}(t; \bm{\mu}) = f(\bm{z}(t; \bm{\mu})) , \quad t\in [0,T]
\end{equation}
where $\bm{z}(t; \bm{\mu}) \in \mathbb{R}^{N_s}$ is defined by Equation \ref{equ:dynamic_state}. 
For a single $\bm{\mu}_i \in \mathcal{D}_{train}$, the associated discrete latent variable matrix $\bm{z}_i$ and the time derivative matrix $\dot{\bm{z}}_i$ can be expressed as
\begin{equation}
    \begin{aligned}
        \bm{z}_i(t; \bm{\mu}) = [z_i^{(1)}(t; \bm{\mu}), z_i^{(2)}(t; \bm{\mu}), \dots, z_i^{(N_s)}(t; \bm{\mu})] \in \mathbb{R}^{N_t+1}\times\mathbb{R}^{N_s}, \\
        \dot{\bm{z}}_i(t; \bm{\mu}) = [\dot{z}_i^{(1)}(t; \bm{\mu}), \dot{z}_i^{(2)}(t; \bm{\mu}), \dots, \dot{z}_i^{(N_s)}(t; \bm{\mu})] \in \mathbb{R}^{N_t+1}\times\mathbb{R}^{N_s},
    \end{aligned}
\end{equation}
where $\dot{\bm{z}}_i$ has typically been computed by finite difference methods or physical priors during the training procedure in previous studies \cite{loiseau2018sparse, 2016SINDy, bonneville2024gplasdi, he2023glasdi, lin2024data}. In this paper,  $\dot{\bm{z}}_i$ is instead obtained through the network $NN_{dyn}$ and iterative cycles of the dynamic Equation \ref{equ:dynamic_state}. 
The form of the governing dynamic function $f$ is determined  to be a linear combination of nonlinear terms from a library of candidate basis functions $\bm{\Theta}(\bm{z}_i)=[b_1(\bm{z}_i), b_2(\bm{z}_i), \dots, b_{N_B}(\bm{z}_i)] \in \mathbb{R}^{N_t+1}\times \mathbb{R}^{N_b}$, defined by the user. $N_b$ is the total number of columns across all basis functions, given by $N_b=\sum_{k=1}^{N_B}N_k$, where $N_k$ denotes the number of columns of $b_k(\bm{z}_i)$.
In other words, Equation \ref{equ:sindy-govern} for each training parameter $\bm{\mu}_i$ is approximated as 
\begin{equation}\label{equ:sindy}
    \dot{\bm{z}_i} \approx \bm{\Theta}(\bm{z}_i) \cdot \bm{\Xi}
\end{equation}
where $\bm{\Xi} = [\bm{\xi}_1, \dots, \bm{\xi}_{N_s}] \in \mathbb{R}^{N_b} \times \mathbb{R}^{N_s}$ is the coefficient matrix. 

To capture the dynamic characteristics of the latent space, the terms in the basis function library $\bm{\Theta}$ can empirically include constants, polynomials, trigonometric functions, exponential functions, etc.. For example, $b_1(\bm{z}_i)$ usually represents a constant term; if $b_{k_1}(\bm{z}_i)$ is a second-order polynomial term, then $N_{k_1}=(N_s+1)N_s/2$ and $b_{k_1}(\bm{z}_i)$ has the specific form:
\begin{equation}
    b_{k_1}(\bm{z}_i) = 
    \begin{pmatrix}
    (z_i^{(1)})^2(t_0) & z_i^{(1)}z_i^{(2)}(t_0) & \cdots & (z_i^{(1)})^2(t_0) & \cdots & (z_i^{(N_s)})^2(t_0) \\\
    (z_i^{(1)})^2(t_1) & z_i^{(1)}z_i^{(2)}(t_1) & \cdots & (z_i^{(1)})^2(t_1) & \cdots & (z_i^{(N_s)})^2(t_1) \\
    \vdots & \vdots & \ddots & \vdots & \ddots & \vdots \\
    (z_i^{(1)})^2(t_{N_t}) & z_i^{(1)}z_i^{(2)}(t_{N_t}) & \cdots & (z_i^{(1)})^2(t_{N_t}) & \cdots & (z_i^{(N_s)})^2(t_{N_t})
    \end{pmatrix}.
\end{equation}

If $b_{k_2}(\bm{z}_i)$ is a cosine term, then $N_{k_2}=N_s$ and $b_{k_2}(\bm{z}_i)$ is of the specific form: 
\begin{equation}
    b_{k_2}(\bm{z}_i) = 
    \begin{pmatrix}
    \cos{z_i^{(1)}(t_0)}  &  \cos{z_i^{(2)}(t_0)}  &  \cdots  &    \cos{z_i^{(N_s)}(t_0)} \\
    \cos{z_i^{(1)}(t_1)}  &  \cos{z_i^{(2)}(t_1)}  &  \cdots  &    \cos{z_i^{(N_s)}(t_1)} \\
    \vdots  & \vdots & \ddots & \vdots  \\
    \cos{z_i^{(1)}(t_{N_t})}  &  \cos{z_i^{(2)}(t_{N_t})}  &  \cdots  &    \cos{z_i^{(N_s)}(t_{N_t})} \\
    \end{pmatrix}
\end{equation}

In this paper, first-order polynomial terms and constant terms are used as components of $\bm{\Theta}$.
For each $\bm{\mu}_i$, there is a specific system of ODEs to identify the latent dynamics. 
In fact, during the physical training and discovery of ID models, evaluating the error between $\bm{z}_i$ and the estimated value $\tilde{\bm{z}}_i$ is inefficient. This is because the right side of Equation \ref{equ:sindy} approximates the time derivative of $\bm{z}_i$ (i.e. $\dot{\bm{z}}_i$), rather than $\bm{z}_i$ itself, which must be obtained by solving ODEs. 
Consequently, the loss function in the ID model must be adjusted to account for the computational error in time differentiation, which can be summarised as 
\begin{equation}
    Loss = ||\frac{d\bm{z}}{dt} - \bm{\Theta}(\bm{z}) \cdot \bm{\Xi}||^2_2.
\end{equation}
This idea was first utilised by Brunton et al in \cite{champion2019data}, in which the errors in the time derivatives of both the original space (i.e. $\frac{d\bm{u}}{dt}$) and the latent space (i.e. $\frac{d\bm{z}}{dt}$)  are factored in the loss function. However,  considering that incorporating terms related to $\frac{d\bm{u}}{dt}$ would significantly increase computational costs and the risk of overfitting, only the loss in $\frac{d\bm{z}}{dt}$ is included in the loss function here. 

For the low-dimensional data related to all training parameters in $\mathcal{D}_{train}$, a joint training approach is employed in TLDNets by the loss function, simultaneously minimising the sum of MSE as follows:
\begin{equation}
    Loss_{dz/dt}(\Xi) =\frac{1}{N_{\bm{\mu}}}\sum_{i=1}^{N_{\bm{\mu}}}\big(\frac{1}{N_s}\sum_{j=1}^{N_s}||\frac{d\bm{z}_i^{(j)}}{dt}-\Xi(\bm{z}_i^{(j)}) \cdot {\xi_i^{(j)}}^T ||_2^2\big),
\end{equation}
where $N_{\bm{\mu}} = |\mathcal{D}_{train}|$. $Loss_{dz/dt}$ is preceded by a loss weight $\omega_{ID}$.




\section{$k$-Nearest Neighbour interpolation with inverse distance weighting}\label{sec:knn}

Based on latent ID models discovered from the training parameter set $\mathcal{D}_{train}$ and the corresponding high-dimensional data $\{\bm{u}_i\}_{\bm{\mu}_i\in \mathcal{D}_{train}}$ , low-dimensional dynamics can be explored at any unseen or trained parameter point $\bm{\mu}$ throughout the parameter space $\mathcal{D}$. To improve the efficiency of parametric study, the $k$-nearest neighbours (KNN) convex interpolation method is used to fit the learned ID model coefficient set $\{\bm{\Xi}_i\}_{\bm{\mu}_i\in \mathcal{D}_{train}}$. 
As an interpolation method with broad application prospects \cite{kramer2013k}, the KNN approach captures the ID model coefficient values of each $\bm{\mu}$ from local features rather than globally, and offers greater accuracy compared to linear interpolation. In contrast to radial basis functions (RBF) and Gaussian process regression (GPR) methods, which ask optimisation to acquire interpolation weights, KNN-based interpolation results in lower computational costs and higher efficiency. 

In this paper, we combine the inverse distance weighting (IDW) function with the KNN method as an interpolation technique. 
This integration ensures the convexity of the fitting interpolation problem and enhances robustness. 
For any testing parameter vector $\bm{\mu}^* \in \mathcal{D}$, denote the set of $k$ nearest sample parameter points to $\bm{\mu}^*$ in $\mathcal{D}_{train}$ as $\mathcal{D}(\bm{\mu}^*)$.  
The interpolation process is performed for each ODE coefficient (denoted as target $y$ in Algorithm \ref{alg:knn-idw} for simplicity) at each parameter point. The mathematical description is as follows: 
\begin{equation}
    \bm{\Xi}^* = \mathcal{I}(\bm{\mu}^*; \{\bm{\Xi}_i\}_{i\in \mathcal{D}(\bm{\mu}^*)}) = \sum_{\bm{\mu}_i\in \mathcal{D}(\bm{\mu}^*)} \varphi_i(\bm{\mu}^*)\bm{\Xi}_i,
\end{equation}
where $\bm{\Xi}^*$ is the ID model coefficient matrix for the test parameter point $\bm{\mu}^*$, and $\varphi_i(\bm{\mu}^*)$ represents the interpolation weights for each trained $\bm{\Xi}_i$. 
In the KNN method, the distance is determined by the Euclidean distance between the parameter points. Specifically, the distance between parameter points $\bm{\mu}_i$ and $\bm{\mu}^*$ is defined as
\begin{equation}\label{equ:distance}
    d(\bm{\mu}_i, \bm{\mu}^*) = \sqrt{\sum_{j=1}^{N_{\bm{\mu}}}(\bm{\mu}_{i,j}-\bm{\mu}^*_j)},
\end{equation}
where $\bm{\mu}_{i,j}$ and $\bm{\mu}_j^*$ are the components of $\bm{\mu}_i$ and $\bm{\mu}^*$ respectively. 
The interpolation weights can be expressed as
\begin{equation}\label{equ:idw}
    \begin{aligned}
    \varphi_i(\bm{\mu}^*) = \frac{d(\bm{\mu}_i,\bm{\mu}^*)^{-p}}{\sum_{\bm{\mu}_j \in \mathcal{D}(\bm{\mu}^*)}d(\bm{\mu}_j, \bm{\mu}^*)^{-p}}, \\[0.2cm]
    \end{aligned}
\end{equation}
in which $p$ is the distance weighting index, generally taken as $p=2$.

Note that the interpolation is actually performed on a per-parameter-point basis and the ODEs coefficients are predicted sequentially rather than all at once, to ensure accuracy. 
If there exists a $\bm{\mu}_i$ such that $d(\bm{\mu}_i,\bm{\mu}^*)\approx 0$, then it is concluded that $\bm{\mu}^*$ is contained in $\mathcal{D}_{train}$, and $\bm{\Xi}_i$ is directly assigned to $\bm{\Xi}^*$.  
The choice of $k$ also affects the final prediction results. Following a preliminary screening process, the value $k$ is determined to be 5. 
 The detailed process of the KNN method can be found in Algorithm \ref{alg:knn-idw}. 



.
\begin{algorithm}[htpb] 
    \renewcommand{\algorithmicrequire}{\textbf{Input:}}
    \renewcommand{\algorithmicensure}{\textbf{Output:}}
    \caption{IDW interpolation based on KNN}
    \label{alg:knn-idw}
    \begin{algorithmic}[1]
    
      \REQUIRE{The training parameter set $\mathcal{D}_{train}$, 
      the single testing parameter point $\bm{\mu}^*$, 
      the $k$ value,  
      and the trained target $\{y_i\}_{\bm{\mu}_i\in \mathcal{D}_{train}}$;}
    
    \ENSURE{The interpolated target $y^*$, 
    $k$ nearest neighbour set $\mathcal{D}(\bm{\mu}^*) \subset \mathcal{D}_{train}$, 
    and the interpolation weights $\{\varphi_i(\bm{\mu}^*)\}_{\bm{\mu}_i\in\mathcal{D}(\bm{\mu}^*)}$.}
    
    \STATE{Calculate distances between every $\bm{\mu}_i \in \mathcal{D}_{train}$ and $\bm{\mu}^*$ by Equation \ref{equ:distance};}
    \STATE{Rank the distance values in ascending order as a list $Dis(\bm{\mu}^*)$;}

    \IF{value 0 is found in the list $Dis(\bm{\mu}^*)$}
        \STATE{Indicate that $\bm{\mu}^*$ lies in $\mathcal{D}_{train}$ and find the index $j$ such that $\bm{\mu}_j \in \mathcal{D}_{train}$ is nearly equivalent to $\bm{\mu}^*$;}
          \STATE{Assign the corresponding $y_j \in \{y_i\}_{\bm{\mu}_i\in \mathcal{D}_{train}}$ to $y^*$ directly;}
    \ELSE
        \STATE{Search for the $k$ nearest parameter points in $\mathcal{D}_{train}$ to $\bm{\mu}^*$ and combine them into a set $\mathcal{D}(\bm{\mu}^*)$;}
        \STATE{Compute the IDW weight $\varphi_i(\bm{\mu}^*)$ by Equation \ref{equ:idw} for $\forall \bm{\mu}_i \in \mathcal{D}(\bm{\mu}^*)$;}
        \STATE{Search for the corresponding $k$ target $y$ based on the parameter points $\bm{\mu} \in \mathcal{D}(\bm{\mu}^*)$ and denote them as $\{y_i\}_{i=1}^k$;}
        \STATE{Compute the target interpolated coefficient $y$ by
        \begin{equation}
            y^{*} = \sum_{i=1}^{k}\varphi_i(\bm{\mu}^*)y_i.
        \end{equation}}
    \ENDIF    
    \RETURN{$y^*$, $\mathcal{D}(\bm{\mu}^*)$, $\{\varphi_i(\bm{\mu}^*)\}_{\bm{\mu}_i\in\mathcal{D}(\bm{\mu}^*)}$.}
    \end{algorithmic}
\end{algorithm}

\section{Data pre-processing}\label{sec:data-preparation}
\vspace{-2pt}

\subsection{Data sampling}

There are various methods for the selection of the training parameter set $\mathcal{D}_{\text{train}}$. To ensure training efficiency, commonly applied methods include greedy sampling based on training results or physical information\cite{he2023glasdi, bonneville2024gplasdi}, uniform sampling\cite{fries2022lasdi}, and Smolyak sparse grid method\cite{xiao2017smolyak, fu2023physics}. The goal of these sampling methods is typically to improve the final performance of algorithms with as few parameter points as possible, especially for those based on autoencoders. This is because, compared to TLDNets,  these algorithms, while also having an excellent ability to capture underlying dynamics, require more network parameters and take up more GPU memory, resulting in longer computation times and the need for more computational resources.

In our work, the random sampling method is used to choose $\mathcal{D}_{train}$. 
For convenience and to observe the performance of P-TLDINets across the entire parameter domain $\mathcal{D}$,  we uniformly sample a set of discrete parameter points $\mathcal{D}^h \subset \mathcal{D}$, with the number of points far exceeding that of $\mathcal{D}_{train}$ (i.e. $|\mathcal{D}^h|>|\mathcal{D}_{train}|)$), to serve as the testing sample set.

\subsection{Normalisation}\label{sec:normalisation}
To enhance the training efficiency of the networks, we preprocess the training data. Especially, there is a need to normalise the inputs and outputs of P-TLDINets. These data include the parameter vector $\bm{\mu}$, the scalar field $\bm{u}$, and the spatial grid coordinate variable $\bm{x}$. Each element is scaled to fit within a specific closed interval, usually chosen as $[-1, 1]$. Based on the data distribution, the normalisation for each variable is conducted independently. The mathematical process is defined as follows:
\begin{equation}
    \overline{\beta} = (\beta-\beta_{ref}) / \beta_{w},
\end{equation}
where $\beta_{ref}$ is a reference value and $\beta_{w}$ is a reference variable range. Particularly, if a variable takes values in the closed interval $[\alpha_{min}, \alpha_{max}]$, then
\begin{equation}\label{equ:alpha-beta}
\left\{
\begin{aligned}
    & \beta_{ref}= (\alpha_{min}+\alpha_{max}) / 2 \\
    & \beta_{w}= (\alpha_{max}-\alpha_{min}) / 2.
\end{aligned}
\right.
\end{equation}
If $\alpha_{min} = -\alpha_{max}$, indicating that the origin is centred in $[\alpha_{min}, \alpha_{max}]$, a range hyper-parameter $\mathcal{L}_{range}$ can be set to adjust the normalisation range as follows:
\begin{equation}\label{equ:alpha-beta-range}
\left\{
\begin{aligned}
    & \beta_{ref}= \mathcal{L}_{range} (\alpha_{min}+\alpha_{max}) / 2 \\
    & \beta_{w}= \mathcal{L}_{range} (\alpha_{max}-\alpha_{min}) / 2.
\end{aligned}
\right.
\end{equation}

After normalisation, the neural network can be expressed as: 
\begin{equation}\label{equ:NN-normalisation}
\left\{
\begin{aligned}
    & \widetilde{NN}_{dyn}(\bm{z}(t), \bm{\mu}) = NN_{dyn}(\bm{z}(t), (\bm{\mu}-\bm{\mu}_{ref})\odot (1/\bm{\mu}_{w}))  \\
    & \widetilde{NN}_{z_0}(\bm{\mu}) = NN_{z_0}((\bm{\mu}-\bm{\mu}_{ref})\odot (1/\bm{\mu}_{w})) \\
    & \widetilde{NN}_{rec}(\bm{z}(t), \bm{\mu}, \bm{x}) = NN_{rec}(\bm{z}(t), \bm{\mu}, (\bm{x}-\bm{x}_0)\odot (1/\bm{x}_0)),
\end{aligned}
\right.
\end{equation}
where $\odot $ denotes the Hadamard product, i.e. element-wise product. This normalisation ensures that the input variables are within a standardised range, thereby facilitating more efficient network training. 
If removing normalisation, the networks can generally still achieve the fitting effect. However, it often requires more epochs of training and computational time.

\begin{figure}[!t]
    \centering
	\includegraphics[width=1.0\linewidth]{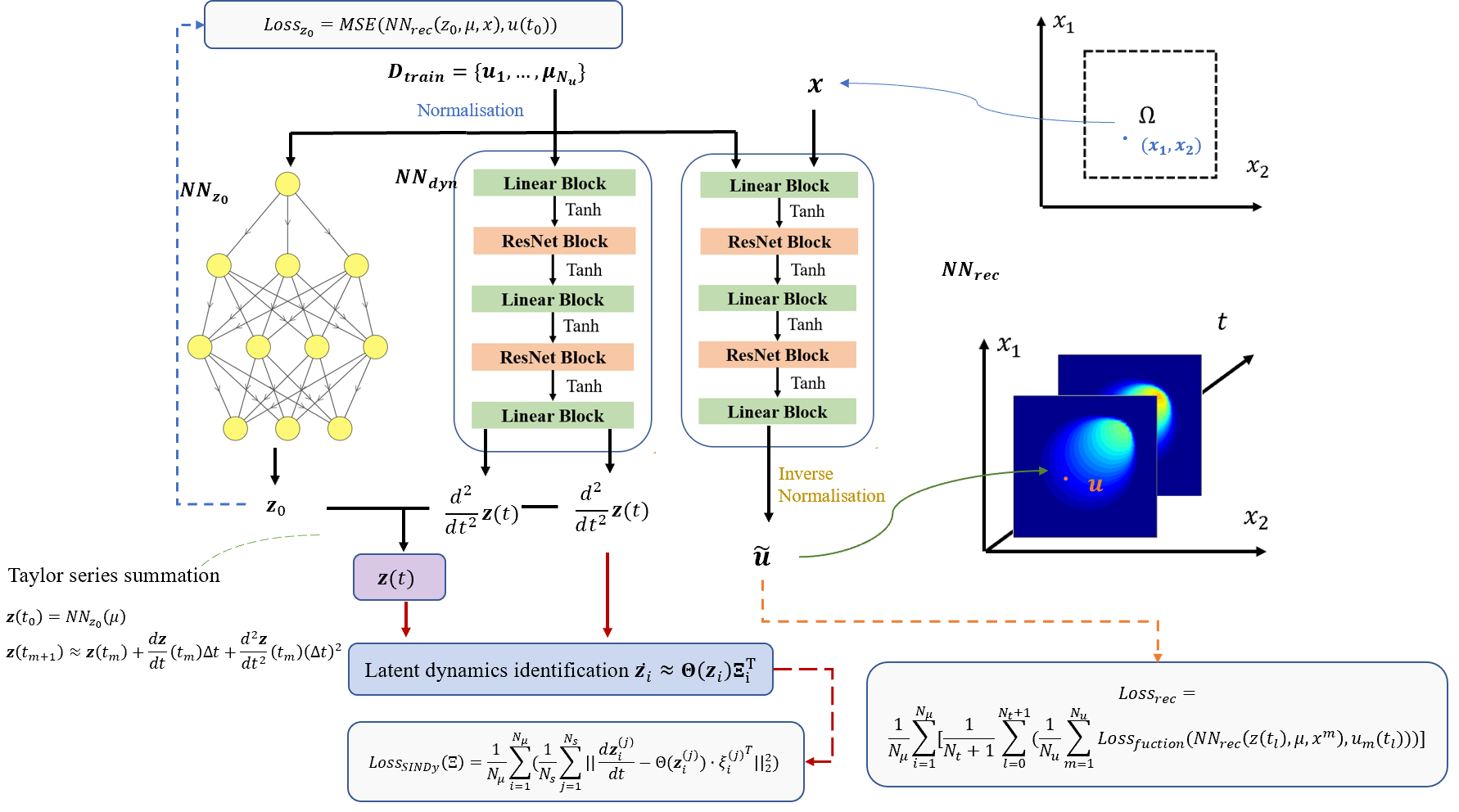}
    \caption{Flowchart of the P-TLDINets algorithm.}
    \label{fig:flow chart}
\end{figure}

\section{Framework of P-TLDINets} \label{sec:framework}

The individual components of P-TLDINets have been described above. 
After preprocessing data, P-TLDINets focus on obtaining a family of ODE coefficients that can be used for parametric learning by training both the TLDNets and the ID model simultaneously. That is achieved by the total loss function as: 
\begin{equation}\label{equ:total-loss}
    Loss_{model} = Loss_{rec}(W_{rec}, W_{dyn}) + \omega_{z_0}Loss_{z_0}(W_{z_0}) + \omega_{ID} Loss_{dz/dt}(\bm{\Xi}) + \omega_{coef} ||\bm{\Xi}||_2^2,
\end{equation}
where the last term $||\bm{\Xi}||_2^2$ is a penalty term. 
This process requires a long training time and is an offline training. 
Then, the KNN interpolation method is used based on the spatial characteristics of the testing parameters, and a series of latent parametric governing equations for each training parameter $\bm{\mu}$ are solved to predict latent state variables $\tilde{\bm{z}}$. The high-dimensional approximate flow field $\tilde{\bm{u}}$ can be quickly simulated using the $NN_{rec}$ model. 
This part is an online process, and it only takes a short time to reach the desired results. 
In addition, trained P-TLDINets allow for direct predictions of $\bm{u}$ based on the inputs of $NN_{rec}$,  which are independent spatial coordinates $\bm{x}^*$ and the testing parameter point $\bm{\mu}^*$. 
The whole process refers to Equation \ref{equ:NN-normalisation}. 




According to the error between the estimated vector $\tilde{\bm{u}}$ and the true value $\bm{u}$, the parametric prediction and spatial prediction effectiveness of the model can be assessed by $L_2$ reconstruction rate as:
\begin{equation}
    r_{L_2}(\tilde{\bm{u}}, \bm{u}) = \frac{{Norm_2}(\tilde{\bm{u}}-\bm{u})}{{Norm_2}(\bm{u})}\times 100 \%, 
\end{equation}
where ${Norm}_2$ stands for $L_2$ norm of matrices. 


A stopping criterion is set for $r_{L_2}(\tilde{\bm{z}}, \bm{z})$ and $Loss_{model}$, which would stop training when they are smaller than hyperparameters $tol_1$ and $tol_2$ set up in advance respectively. 


\subsection{Off-line model training}

The details mentioned in the previous sections together form the P-TLDINets algorithm based on parametric learning. In practice, after randomly selecting a parameter points subset $\mathcal{D}_{train} \subset \mathcal{D}$, a small amount of high-fidelity snapshot is generated by simulation techniques as training data. 
These data are then fed into the P-TLDINets model after going through preprocessing, especially the normalisation process in Section \ref{sec:normalisation}. A series of latent governing equations with the form of Equation \ref{equ:sindy-govern} are identified to describe the intrinsic dynamics. 
The coefficient matrices $\{\bm{\Xi}_i\}$ of Equation \ref{equ:sindy-govern} for each $\bm{\mu}_i \in \mathcal{D}_{train}$ , treated as trainable variables within the networks, are derived and compiled into a set for subsequent online computations as a significant ingredient of the parametric prediction. 
The specific procedure can be referred to Algorithm \ref{alg:train-model} and the flow chart is in Figure \ref{fig:flow chart}. 

\begin{algorithm}[htpb] 
    \renewcommand{\algorithmicrequire}{\textbf{Input:}}
    \renewcommand{\algorithmicensure}{\textbf{Output:}}
    \caption{P-TLDINets trained with random sampling}
    \label{alg:train-model}
    \begin{algorithmic}[1]
    
    \REQUIRE{The training parameter set $\mathcal{D}_{train}$, 
    the high-dimensional simulated field data $\{\bm{u}_i\}_{\bm{\mu}_i\in\mathcal{D}_{train}}$, 
    the spatial coordinates information $\bm{x}$, 
    the discrete time $t$, 
    the max iteration $iter$, 
    the target tolerance $tol_1$ and $tol_2$, 
    the number of epoch to test the tolerance $N_{tol}$, 
    the weight $\omega_{ID}$ of $Loss_{dz/dt}$, 
    the weight $\omega_{z_0}$ of $Loss_{z_0}$, 
    the weight $\omega_{coef}$ of $Loss_{coef}$;}
    
    \ENSURE{The model coefficient $W_{rec}$, $W_{dyn}$ and $W_{z_0}$, 
    the interpolated coefficient matrices $\{\bm{\Xi}_i\}_{\bm{\mu}_i\in\mathcal{D}_{train}}$.}

    \STATE{Preprocess $\bm{\mu}_i \in \mathcal{D}_{train}$, $\{\bm{u}_i\}_{\bm{\mu}_i\in\mathcal{D}_{train}}$  and $\bm{x}$ by normalisation in Section \ref{sec:normalisation};}

    \STATE{Construct P-TLDINets model according to the nature of problem and fed it with normalised $\bm{\mu}_i$, $\bm{u}_i$ and $\bm{x}$;}

    \WHILE{$i \leq iter$}
        \STATE{Update $W_{rec}$, $W_{dyn}$ and $W_{z_0}$ by minimising $Loss_{model}$ in Equation \ref{equ:total-loss};}
        \IF{$i  \%  N_{tol} = 0$}
            \STATE{Solve the latent governing equations for each training parameter $\bm{\mu}_i$ as
            \begin{equation}
                \frac{d\bm{z}_i}{dt} = \bm{\Theta(\bm{z}_i)} \cdot \bm{\Xi}_i, \quad \bm{\mu}_i\in\mathcal{D}_{train}
            \end{equation}
            to obtain $\{\tilde{\bm{z}}_i\}_{\bm{\mu}_i\in\mathcal{D}_{train}}$ and then compute the $L_2$ error $r_{L_2}$ between $\tilde{\bm{z}}_i$ and $\bm{z}_i$;}
            \IF{$r_{L_2}\leq tol_1$ and $Loss_{model}\leq tol_2$}
            \STATE{\textbf{break}}
            \ENDIF
        \ENDIF
    \ENDWHILE
    
    \RETURN{ $W_{rec}$, $W_{dyn}$, $W_{z_0}$,  $\{\bm{\Xi}_i\}_{\bm{\mu}_i\in\mathcal{D}_{train}}$.}

    \end{algorithmic}
\end{algorithm}

\begin{figure}[!t]
    \centering
	\includegraphics[width=1.0\linewidth]{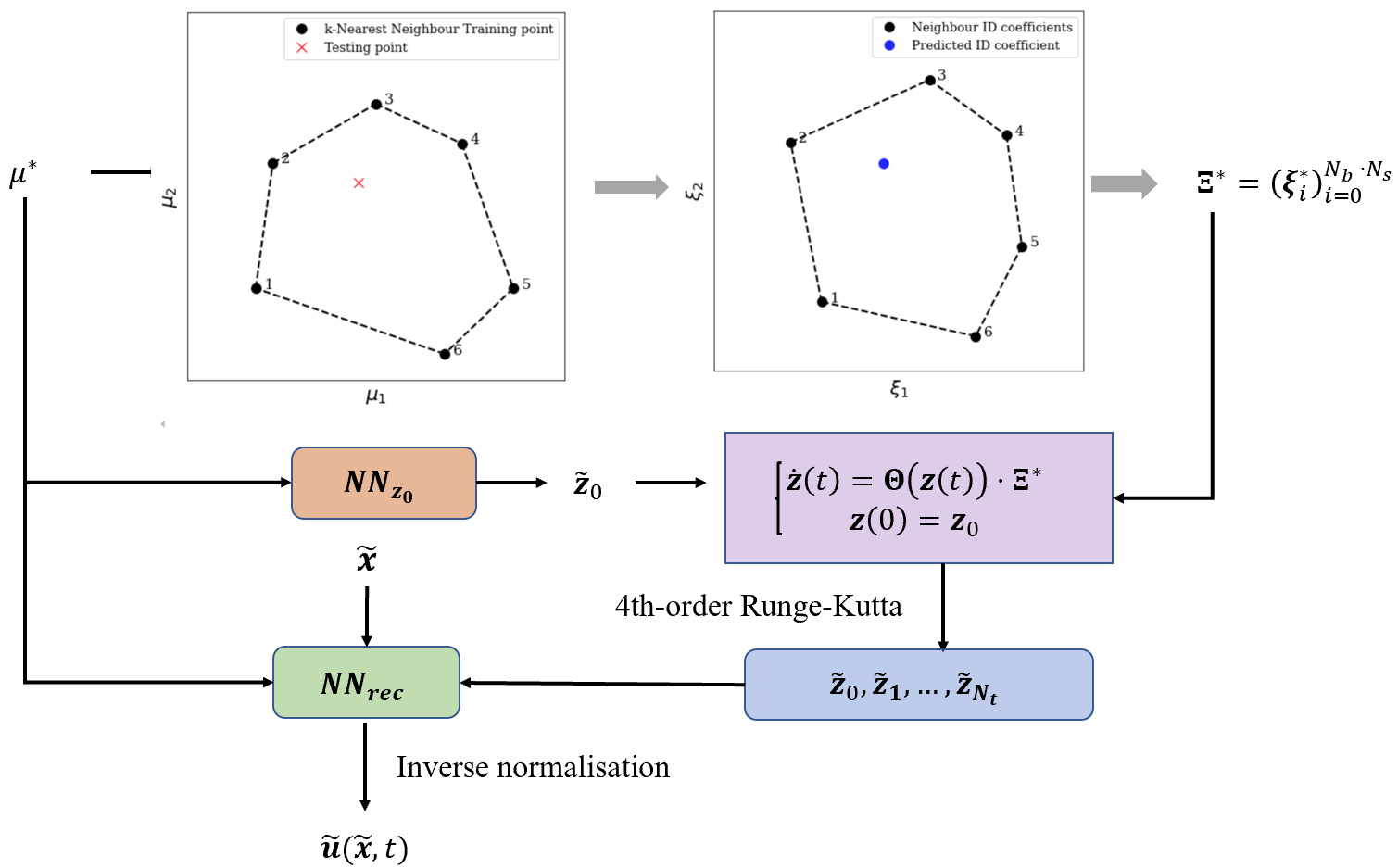}
    \caption{Online stage of the P-TLDINets algorithm.}
    \label{fig:online chart}
\end{figure}

\subsection{Online stage with the reconstruction model and ID models }


The trained P-TLDINets model is then employed in efficient predictions of estimated solutions for a given parameter. During the online stage, the prediction for unknown parameter points, or even for spatial grid coordinates that were not included or partly missed in the training data, is able to be accomplished with reduced time and costs. 
By combining $\bm{\mu}^*$  and $\bm{x}^*$ information with the set of coefficients $\{\bm{\Xi}_i\}_{\bm{\mu}_i\in\mathcal{D}_{train}}$, high-dimensional estimates is obtained through KNN convex interpolation and the trained network $NN_{rec}$ and $NN_{z_0}$. 
The detailed process is shown in Algorithm \ref{alg:online-line-predict}

\begin{algorithm}[htbp]
    \renewcommand{\algorithmicrequire}{\textbf{Input:}}
    \renewcommand{\algorithmicensure}{\textbf{Output:}}
    \caption{Evaluation and online stage of the P-TLDINets model}
    \label{alg:online-line-predict}
    \begin{algorithmic}[1]
    
    \REQUIRE{A testing parameter set $\mathcal{D}^h$, 
    the training parameter set $\mathcal{D}_{train}$, 
    the trained reconstruction model coefficient $W_{rec}$, 
    the trained ODE coefficient matrices $\{\bm{\Xi}_i\}_{\bm{\mu}_i\in \mathcal{D}_{train}}$, 
    spatial coordinate information $\bm{x}^*$ to be predicted ;}
    
    \ENSURE{The approximate fields $\{\bm{u}(\bm{\mu}_j)\}_{\bm{\mu}_j\in\mathcal{D}^h}$.}
    \FOR{$\forall \bm{\mu}^* \in \mathcal{D}^h$}
        \FOR{the $m$-th coefficient in ID model coefficient matrix $\bm{\Xi}$ in Equation \ref{equ:sindy}}
            \STATE{Search for the corresponding $k$ target coefficients based on the parameter points $\bm{\mu} \in \mathcal{D}(\bm{\mu}^*)$ and denote them as $\{\xi_i^{(m)}\}_{i\in\mathcal{D}(\bm{\mu}^*)}$;}
            \STATE{Compute KNN weights by Algorithm \ref{alg:knn-idw} and calculate the $m$-th interpolated coefficient $\xi^{*(m)}$ by
            \begin{equation}
                \xi^{*(m)} = \sum_{\bm{\mu}_i\in\mathcal{D}(
                \bm{\mu}^*)}\varphi_i(\bm{\mu}^*)\xi_i^{(m)};
            \end{equation}}
        \ENDFOR
        \STATE{Compute $\bm{\Xi^*} = \{\xi^{*(m)}\}_{m=1}^{N_s \cdot N_b}$ based on the above calculations;}
        \STATE{Solve Equation \ref{equ:sindy-govern} by Runge-Kutta method and obtain $\{\tilde{\bm{z}}_j\}_{\bm{\mu}_j\in\mathcal{D}^h}$;}
        \STATE{Reconstruct approximation scalar data $\bm{u}(\bm{\mu}_j)$ by the reconstruction model $NN_{rec}$ , $\bm{x}^*$ and inverse normalisation as 
        \begin{equation}
            \tilde{\bm{u}}(\bm{x}^*, t; \bm{\mu}_j) = NN_{rec}(\tilde{\bm{z}}, \bm{\mu}, \bm{x}^*)
        \end{equation}.
        }
    \ENDFOR
    \RETURN{$\{\bm{u}(\bm{\mu}_j)\}_{\bm{\mu}_j\in\mathcal{D}^h}$.}

    \end{algorithmic}
\end{algorithm}

\section{Numerical Results}\label{sec:results}
\vspace{-2pt}

This section demonstrates the application of P-TLDINets. 
We explore P-TLDINets on two nemerical examples, the 2D-Burgers equation and lock exchange. 
Within the selected parameter range, the latter example can standardise initial values through certain preprocessing, while the former one cannot. The Burgers equation uses a uniform grid, whereas the lock-exchange problem uses an unstructured triangular mesh, ensuring that both are capable of high-fidelity numerical solutions. We depict the prediction performance and training speed of P-TLDINets with unknown testing parameters. 

In the results presentation of the first example, this method is compared with some dynamics identification methods based on autoencoders. Additionally, two parameter ranges are used to demonstrate the generalisation ability of this framework.
To illustrate the capability of P-TLDINets to handle different coordinates or meshes, we also added a subsection to these two numerical cases, where different scales of meshes (multiscale meshes) are fed into the model.

The training of P-TLDINets and the compared methods are all performed on a NVIDIA GeForce RTX 4090 (Ada Lovelace) with 24GB GDDR6X GPU. 

\subsection{Case 1: 2D-Burgers equation}\label{sec:2d-burgers}

We consider a 2D parametric Burgers equation with a viscous diffusion term, which involves a governing equation and an essential boundary condition. It can be summarised as  
\begin{equation}\label{equ:2D-Burgers}
    \begin{aligned}
        \frac{\partial \bm{u}}{\partial t} + \bm{u} \cdot \nabla \bm{u} & = \frac{1}{Re} \Delta \bm{u}, \quad (x_1, x_2) \in \Omega, t \in [0, 1]  \\[0.1cm]
        \bm{u}(x_1, x_2, t=0; \bm{\mu}) & = 0,  \quad (x_1, x_2) \in \partial \Omega
    \end{aligned}
\end{equation}
where $\Omega = [-3, 3] \times [-3, 3]$ and the Reynolds number $Re = 10000$. 
The symbol $\bm{u}=(u,v)$ represents a vector field, where $u$ and $v$ are the fluid velocity components in the $x$ and $y$ directions, respectively. 
The initial condition determines the evolution of Equation \ref{equ:2D-Burgers}. In this case, it is parameterised by $\bm{\mu} = \{a, \omega \}$ and defined as 
\begin{equation}\label{equ:burgers-initial}
    \bm{u}(t=0,x_1,x_2) = a \exp{(-\frac{x_1^2+x_2^2}{2\omega^2})},
\end{equation}
causing the initial values to stay unequal even after normalisation. 
The high-fidelity solver uses a backward finite-difference discretisation in space and an implicit backward Euler integration scheme in time. 
In the domain $\Omega$, a spatial step size of 50 is used for each edge and a temporal step number of 200 is used, resulting in $\Delta x_1 = \Delta x_2 = 0.12$ and $\Delta t=5\times 10^{-3}$. 
\begin{figure}[htbp]
    \centering
	\includegraphics[width=0.3\linewidth]{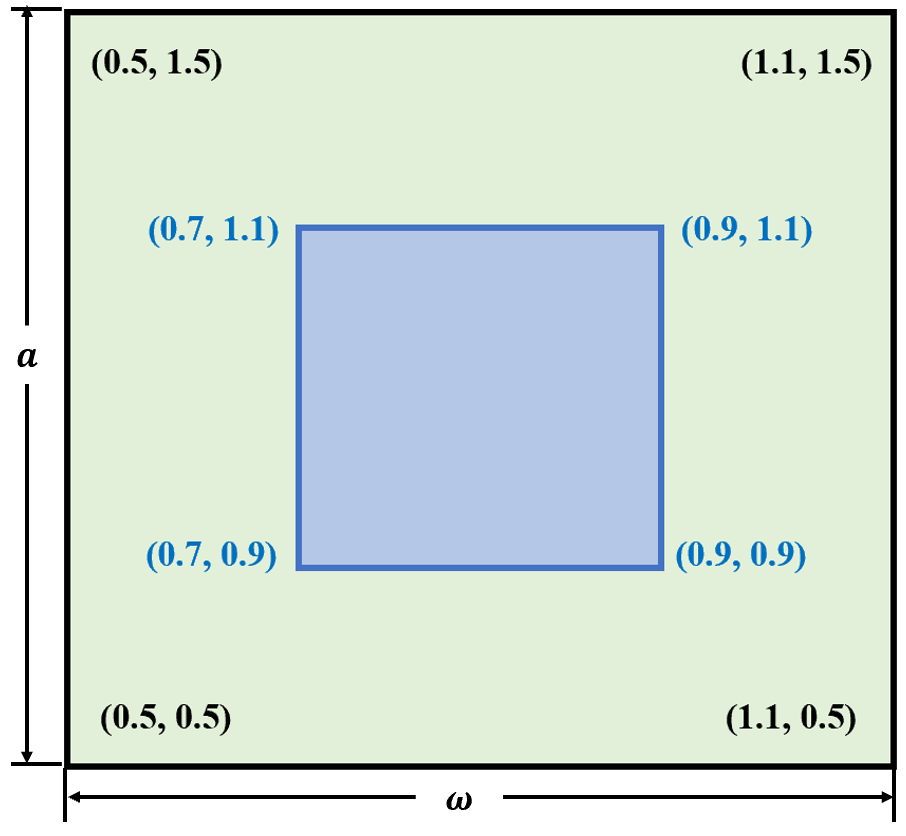}
    \caption{\textbf{2D-Burgers}:  The inner small blue square represents the range of the small parameter set; the outer large green square represents the range of a large parameter set (includes the blue square area).}
    \label{fig:burgers-para-range}
\end{figure}

To demonstrate the generality of the P-TLDINets method, we choose two parameter ranges for comparison, namely a small dataset $\mathcal{D}_1 = [0.7, 0.9] \times [0.9, 1.1]$ and a large dataset $\mathcal{D}_2=[0.5, 1.1] \times [0.5, 1.5]$, as shown in detail in Figure \ref{fig:burgers-para-range}. 

\subsubsection{Trained on a small dataset $\mathcal{D}_1$}


\begin{figure}[!htbp]
\centering
\begin{tabular}{cc}
\begin{minipage}{0.9\linewidth}
\includegraphics[width = \linewidth,angle=0,clip=true]{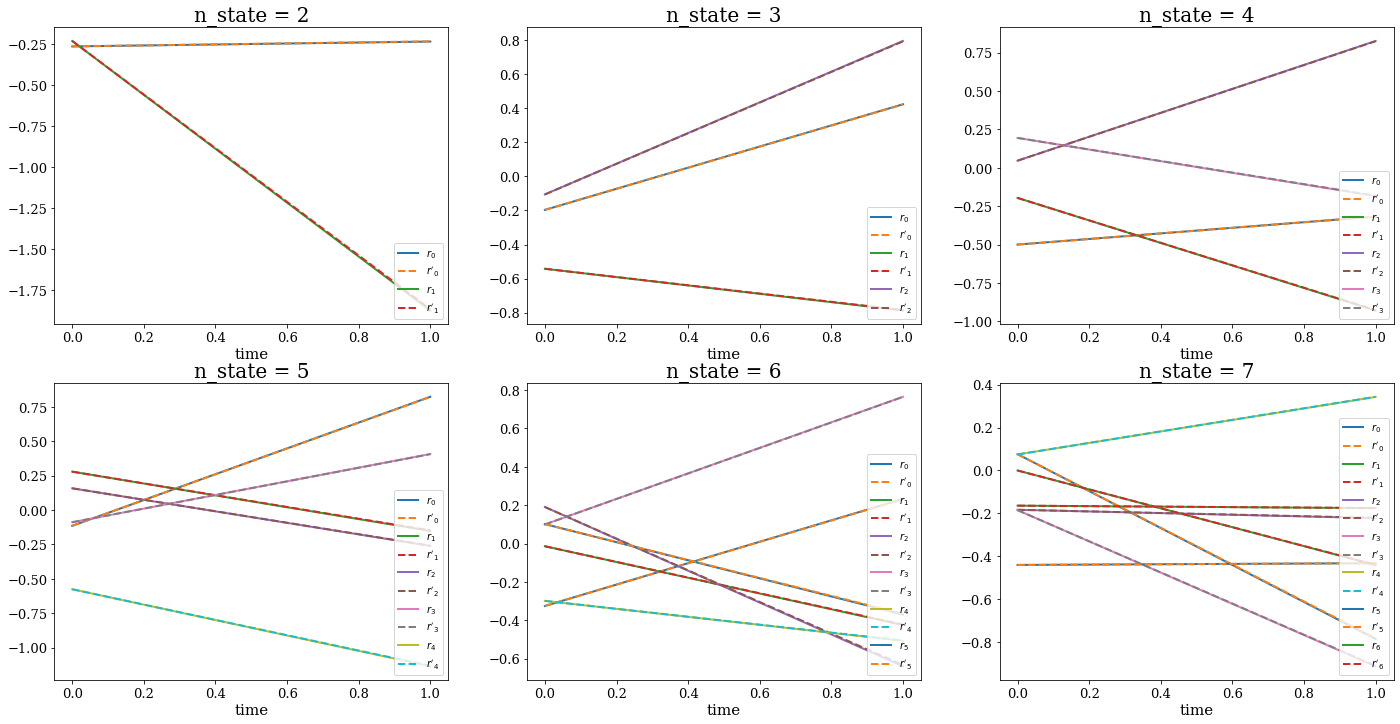}  
\end{minipage}
\\
(a) {\small }
\\
\begin{minipage}{0.9\linewidth}
\includegraphics[width = \linewidth,angle=0,clip=true]{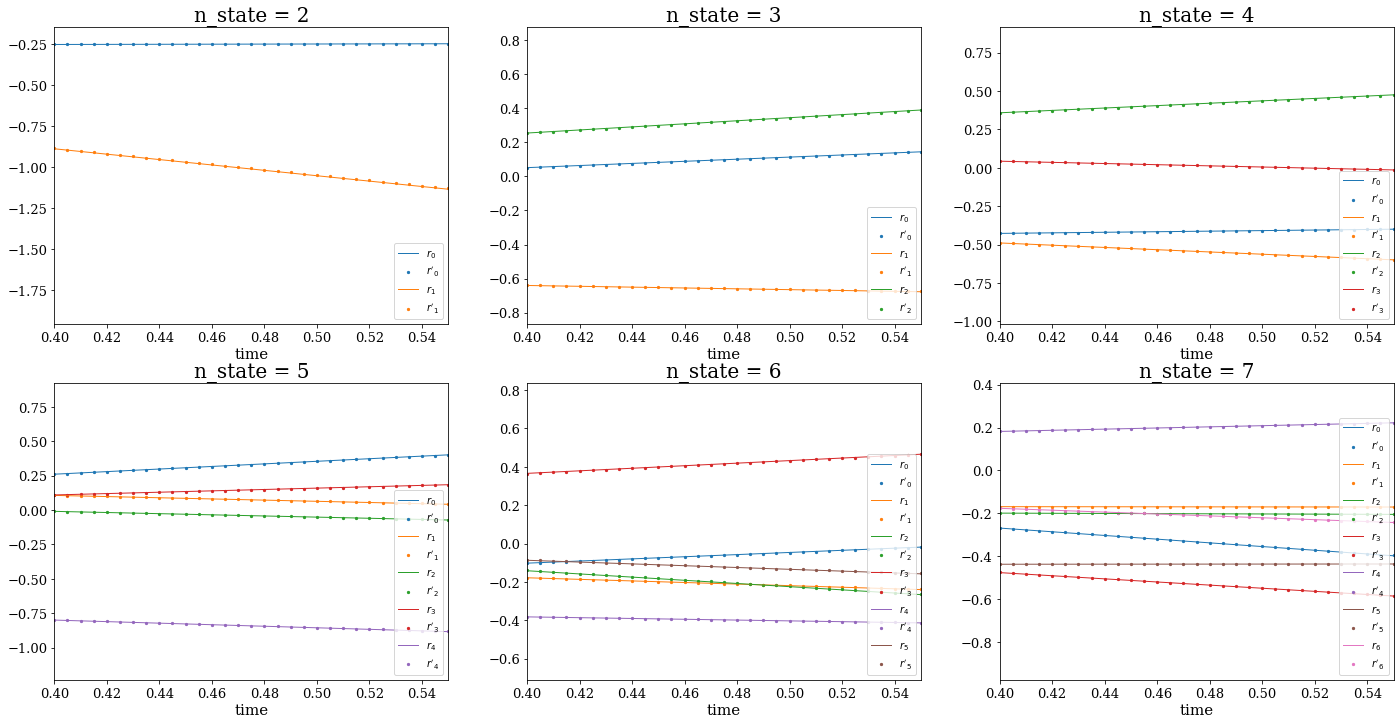}  
\end{minipage} 
 \\
(b) {\small }
\\
\end{tabular}
\caption{\textbf{2D-burgers}: Low-dimensional dynamics and the solutions of the identified equations in the small dataset $\mathcal{D}_1$. (a) and (b) show the dimension of the latent space $N_s$ that ranges from $2$ to $7$. (a) represents the results for the entire time frame $t \in [0, 1.0]$, while (b) shows only in $t \in [0.4, 0.6]$.}
\label{fig:burgers-latent-small}
\end{figure} 

In the first example, P-TLDINets are trained on a small data set $\mathcal{D}_1$, when the dimensionality of the latent space $N_s$ ranges from 2 to 7 to reveal some special features of the reduction mapping in TLDNets. 
The input size of the dynamical network $NN_{dyn}$ is $N_s+N_D=N_s+2$ and the output size is $2N_s$. The specific structure of $NN_{dyn}$ is:
\begin{equation}
    \begin{aligned}
        NN_{dyn}: & \rm{Linear}(N_s+2,6) 
    \stackrel{\rm{Tanh}}{\longrightarrow} \rm{ResNetBlock}(6, 6) 
    \stackrel{\rm{Tanh}}{\longrightarrow} \rm{Linear}(6,6) 
    \stackrel{\rm{Tanh}}{\longrightarrow} \\
    & \rm{ResNetBlock}(6, 6)
    \stackrel{\rm{Tanh}}{\longrightarrow} \rm{Linear}(6,2N_s),
    \end{aligned}
\end{equation}
where "$\rm{Linear}$" is a FCNN and "$\rm{ResNetBlock}$" refers to a ResNet block as mentioned in Section \ref{sec:resnet}. 
The input size of the reconstruction network $NN_{rec}$ is $N_S+2+N_D=N_s+4$ ($2$ denotes the dimension of the spatial domain $\Omega$), and the output size is the number of flow field in $\bm{u}$, which is $2$ in this problem. The specific structure of $NN_{rec}$ is:
\begin{equation}
    \begin{aligned}
        NN_{rec}: & \rm{Linear}(N_s+4,11) 
    \stackrel{\rm{Tanh}}{\longrightarrow} \rm{ResNetBlock}(11, 15) 
    \stackrel{\rm{Tanh}}{\longrightarrow} \rm{Linear}(15,15)
    \stackrel{\rm{Tanh}}{\longrightarrow}   \\
    & \rm{ResNetBlock}(15, 15)
    \stackrel{\rm{Tanh}}{\longrightarrow} \rm{Linear}(15,15)
    \stackrel{\rm{Tanh}}{\longrightarrow} \rm{ResNetBlock}(15, 11)
    \stackrel{\rm{Tanh}}{\longrightarrow}  \\
    & \rm{Linear}(11,2).
    \end{aligned}
\end{equation}
The initial mapping network $NN_{z_0}$ is expressed as: 
\begin{equation}
    \begin{aligned}
        NN_{z_0}: \rm{Linear}(2,6) 
    \stackrel{\rm{Tanh}}{\longrightarrow} \rm{Linear}(6, 6) 
    \stackrel{\rm{Tanh}}{\longrightarrow} \rm{Linear}(6,6)
    \stackrel{\rm{Tanh}}{\longrightarrow} \rm{Linear}(6,N_s).
    \end{aligned}
\end{equation}

To identify dynamics in the latent space $\mathcal{S}$, linear terms and constant comprise the library $\bm{\Theta}(\cdot)$ in Equation \ref{equ:sindy-govern}, leading to $N_b=6$.  The coefficient matrix $\bm{\Xi}$ is contained in the network as trainable variables. 
Compared to the autoencoder, this is a sufficiently lightweight structure, with fewer model trainable parameters, leading to more rapid model training and less use of random access unit. 

The P-TLDINets model is trained by the Adam optimiser for 6000 iterations, with $N_{tol}=500$. To further refine the training, a learning rate scheduler (StepLR) is implemented with an initial learning rate $\alpha = 0.05$, which decreased $\alpha$ by a factor of 0.6 every 500 iterations. 
25 samples are randomly selected to estimate the latent dynamics. The loss weights are set as $\omega_{DI}=0.05$, $\omega_{coef}=0$, $\omega_{z_0}=0.5$.

\begin{figure}[htbp]
\centering
\begin{tabular}{cc}
\begin{minipage}{0.7\linewidth}
\includegraphics[width = \linewidth,angle=0,clip=true]{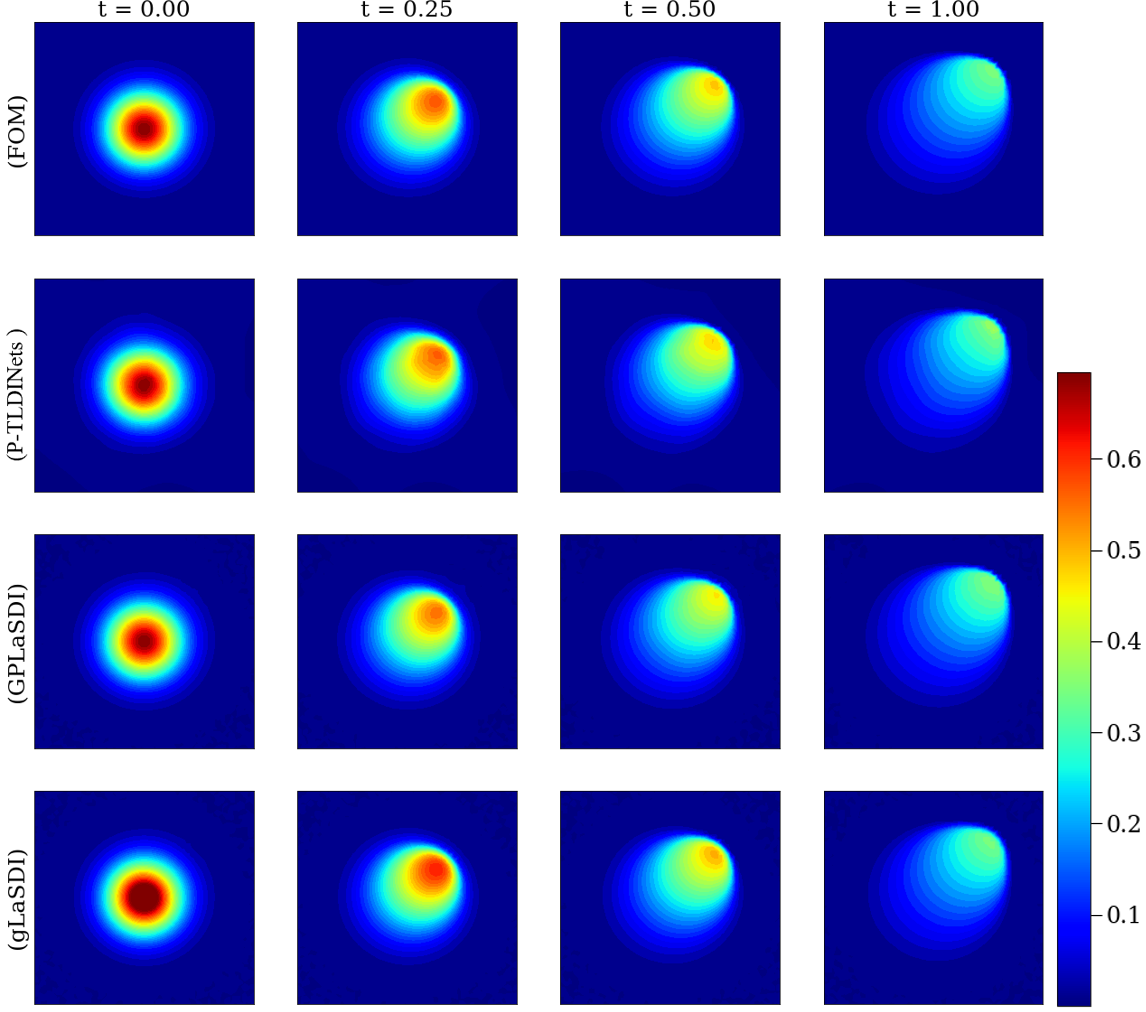}  
\end{minipage}
\\
(a) {\small Solutions}
\\
\begin{minipage}{0.7\linewidth}
\includegraphics[width = \linewidth,angle=0,clip=true]{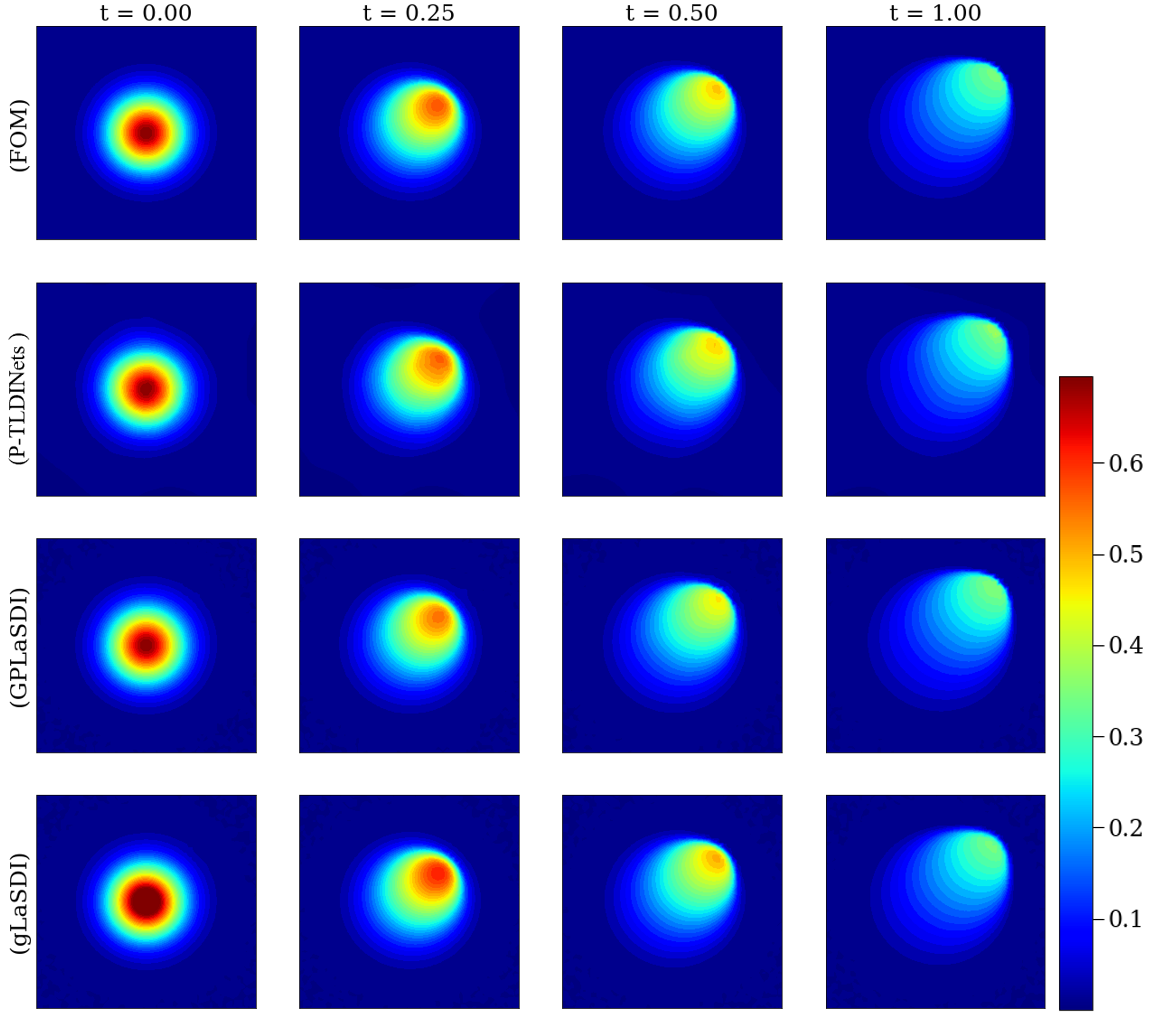}  
\end{minipage} 
\\
(b) {\small Errors}
\\
\end{tabular}
\caption{\textbf{2D-burgers}: (a) shows high-fidelity velocity solutions and reconstruction results of P-TLDINets, gLaSDI and GPLaSDI on $\bm{\mu}^*=[0.71, 1.04]$ in $\mathcal{D}_1^h$ when $N_s=5$. (b) demonstrates point-wise errors between three types of methods and the true value. }
\label{fig:burgers-solution-error-small}
\end{figure}

\begin{table}[h!]
\centering
\label{table:burgers-error-dimenion-small}
\begin{tabular}{c|cccccc}
   \toprule
   $N_s$ & 2 & 3 & 4 & 5 & 6 & 7 \\
   \midrule
   $L_2$ error ($\%$)  & 0.7504  &  0.7010 & 0.7531  & 0.6623  &  0.8467 &  0.7951 \\
   \bottomrule
\end{tabular}
\caption{\textbf{2D-Burgers}: Total $L_2$ errors on testing samples $\mathcal{D}^h \subset \mathcal{D}_1$ of different $N_s$, which are the dimension of the latent dynamics.}
\end{table}

The results of different $N_s$ are shown in Figure \ref{fig:burgers-latent-small}. To evaluate the reconstruction performance of $NN_{rec}$ and $\{\bm{\Xi}_i\}_{\bm{\mu}_i\in\mathcal{D}_1}$ with the KNN interpolation method, 225 test samples in $\mathcal{D}_1$ are adopted as $\mathcal{D}_1^h$. The overall $L_2$ reconstruction errors are shown in Table \ref{table:burgers-error-dimenion-small}. It can be seen that P-TLDINets still demonstrate sufficient compression and learning capabilities even with different $N_s$. 
It takes 1.3 hours to obtain high-dimensional solutions for the entire testing set $\mathcal{D}^h_1$ for the solver, whereas P-TLDINets only require 36.35 minutes of training time and 57.02 seconds to predict on $\mathcal{D}^h_1$, achieving a striking acceleration of $94.7\times$ speed-up. From Table \ref{table:burgers-error-dimenion-small}, P-TLDINets reach the reconstruction accuracy of $99\%$ on the test dataset for every latent dimension. 

\begin{figure}[!ht]
    \centering
	\includegraphics[width=0.6\linewidth]{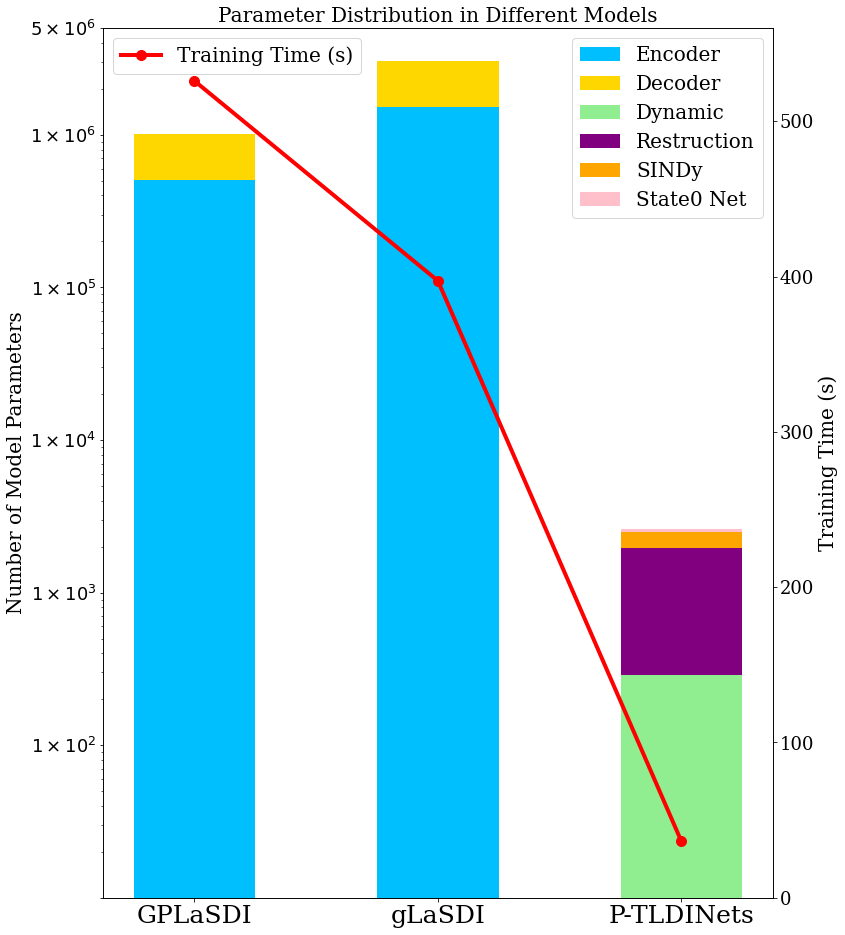}
    \caption{\textbf{2D-Burgers}: 
    The bar chart shows the trainable model parameters and their compositions for GPLaSDI, gLaSDI, and P-TLDINets; the red line indicates the required model training time. 
    GPLaSDI and gLaSDI consist of an encoder and a decoder, whereas P-TLDINets replace these with a combination of a dynamic network, a reconstruction network, and $z_0$ network.} 
    \label{fig:burgers-param-time-comparision}
\end{figure}

In Figure \ref{fig:burgers-solution-error-small}, we compare the model with advanced reduced-order algorithms, GPLaSDI\cite{bonneville2024gplasdi} and gLaSDI\cite{he2023glasdi}, which are based on autoencoders and identification of dynamics. We pick $\bm{\mu}^*=[0.71, 1.04]$ randomly to show the predicted solutions of different methods at several time points when $N_s=5$. Since $u$ and $v$ in $\bm{u}$ are highly similar in the initial condition of Equation \ref{equ:burgers-initial}, only the variation of $u$ is shown here.
All three methods roughly restore the trends of the flow fields, and Figure \ref{fig:burgers-solution-error-small}(b) presents that their errors remain within a very narrow range.
However, as shown in Figure \ref{fig:burgers-param-time-comparision}, gLaSDI and GPLaSDI contain far more trainable parameters and require more training time compared to P-TLDINets. To further demonstrate the effects of P-TLDINets, a large parameter dataset $\mathcal{D}_2$ is employed in Section \ref{sec:burgers-d2}. 

\subsubsection{Trained on a large dataset $\mathcal{D}_2$}\label{sec:burgers-d2}


\begin{figure}[!t]
\centering
\begin{tabular}{cc}
\begin{minipage}{0.9\linewidth}
\includegraphics[width = \linewidth,angle=0,clip=true]{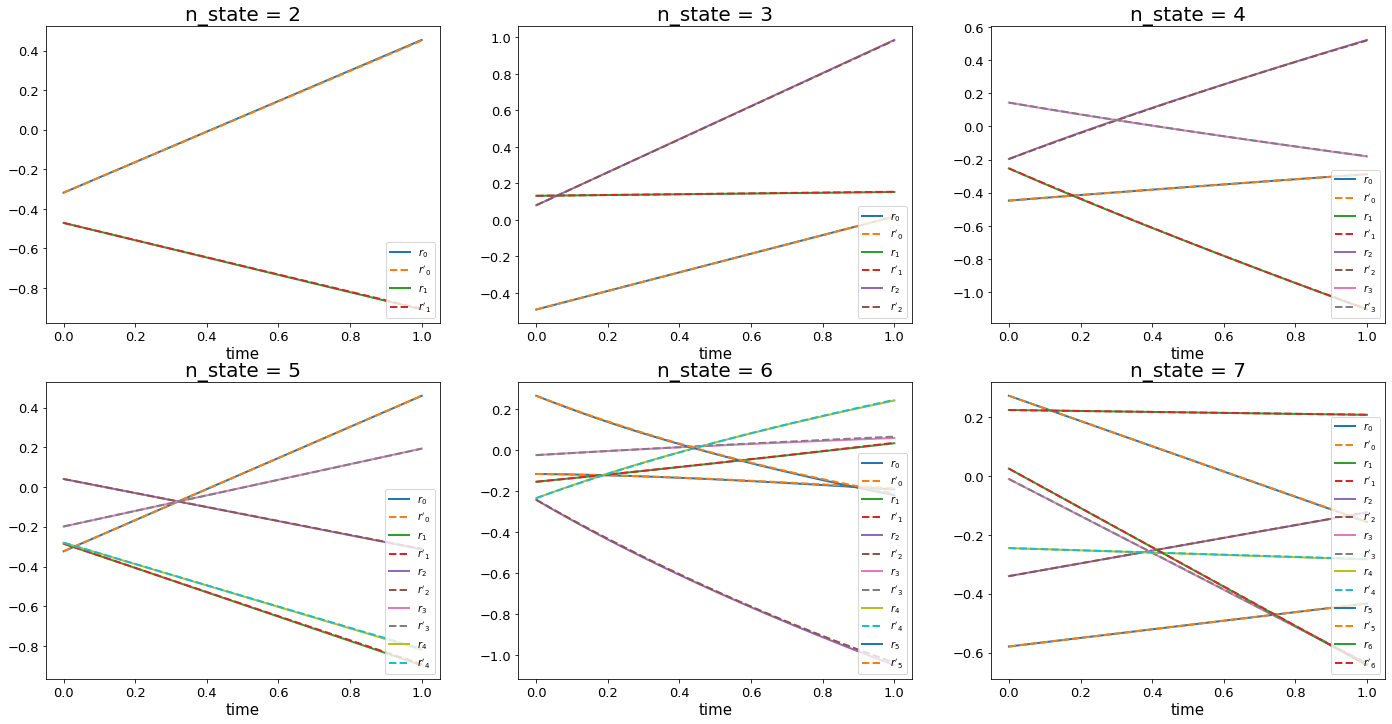}  
\end{minipage}
\\
(a) {\small }
\\
\begin{minipage}{0.9\linewidth}
\includegraphics[width = \linewidth,angle=0,clip=true]{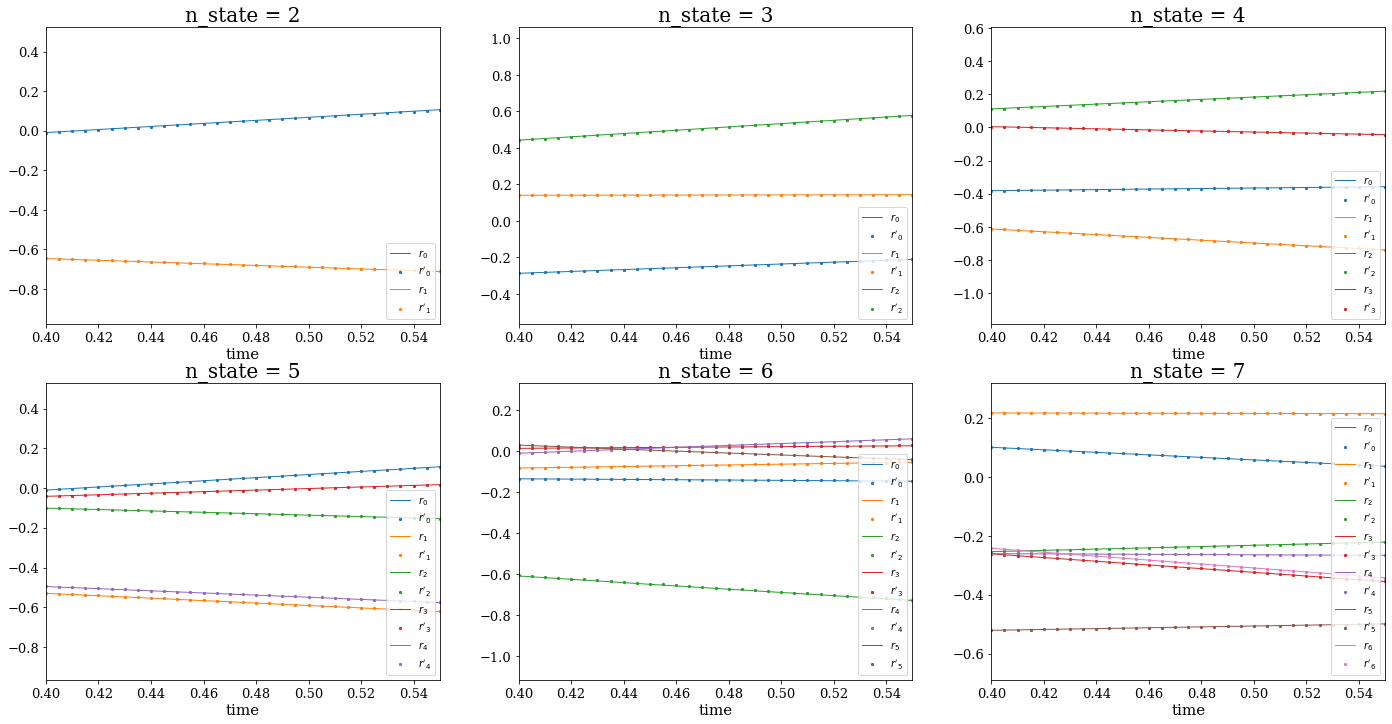}  
\end{minipage} 
\\
(b) {\small }
\\
\end{tabular}
\caption{\textbf{2D-burgers}: Low-dimensional dynamics and the solutions of the identified equations on the large dataset $\mathcal{D}_2$. (a) and (b) show the dimension of latent space $N_s$ ranging from $2$ to $7$. (a) represents results for the entire time frame $t \in [0, 1.0]$, while (b) shows only in $t \in [0.4, 0.6]$.}
\label{fig:burgers-latent-new}
\end{figure} 

For Burgers equation, a larger range of parameters implies more complex dynamics.  
The network structure and training information are the same as those in the previous section. A new testing parameter point set, $\mathcal{D}^h_2$, also containing 225 sample parameters, is selected from $\mathcal{D}_2$. 

\begin{figure}[htbp]
\centering
\begin{tabular}{cc}
\begin{minipage}{0.7\linewidth}
\includegraphics[width = \linewidth,angle=0,clip=true]{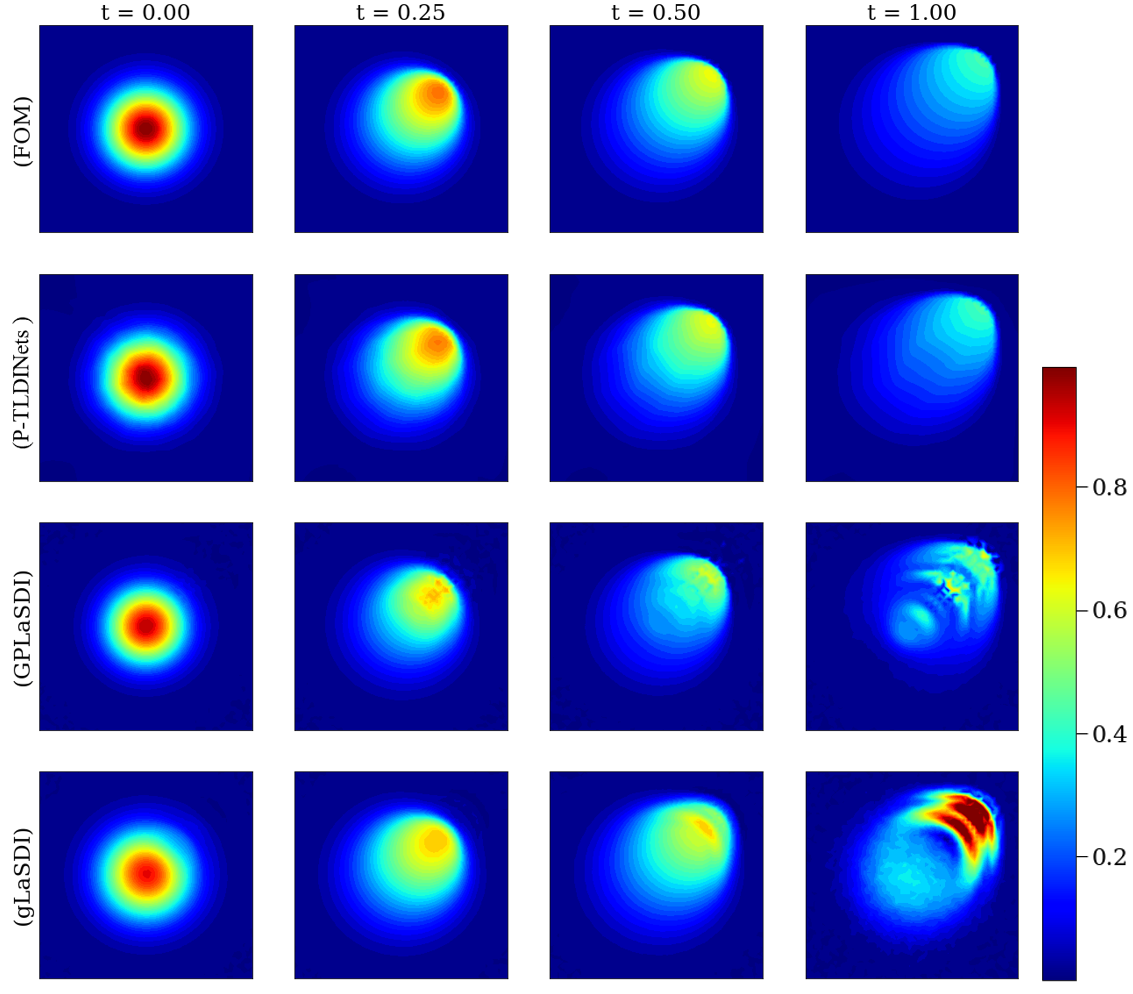}  
\end{minipage}
\\
(a) {\small velocity solutions}
\\
\begin{minipage}{0.7\linewidth}
\includegraphics[width = \linewidth,angle=0,clip=true]{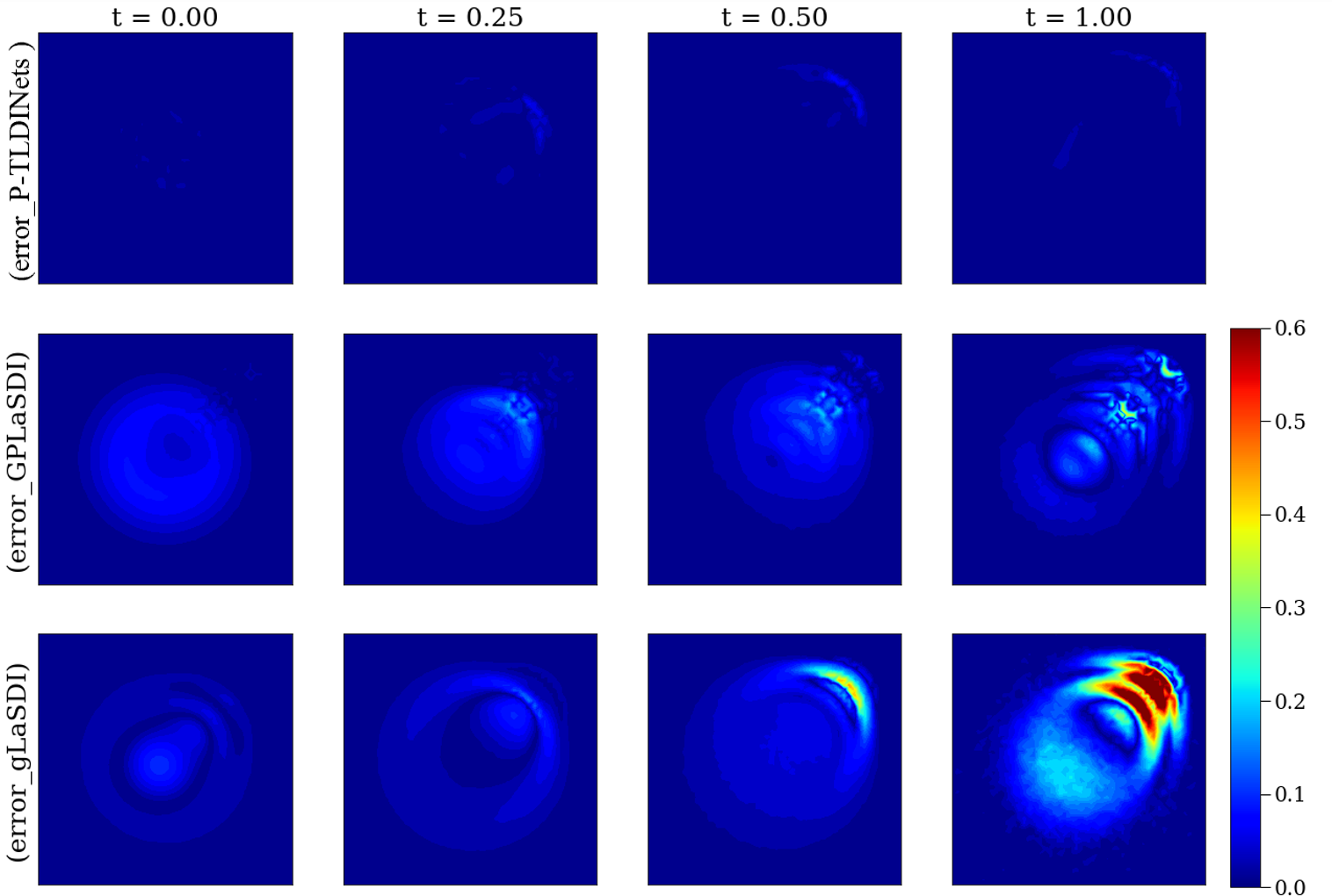}  
\end{minipage} 
\\
(b) {\small Errors} 
\\
\end{tabular}
\caption{\textbf{2D-Burgers}: (a) shows high-fidelity velocity solutions and reconstruction results of P-TLDINets, gLaSDI and GPLaSDI on $\bm{\mu}^*=[1.0, 1.29]$ in $\mathcal{D}_2^h$ when $N_s=5$. (b) demonstrates point-wise errors between three types of methods and true value.}
\label{fig:burgers-solution-error-new}
\end{figure}

Figure \ref{fig:burgers-latent-new} illustrates the low-dimensional temporal dynamics of P-TLDNets for $N_s$ ranging from 2 to 7 as well. Figure \ref{fig:burgers-solution-error-new} presents the prediction results of GPLaSDI, gLaSDI and P-TLDINets for the testing parameter $\bm{\mu}=[1.0, 1.29]$ when $N_s=5$. 
In $\mathcal{D}_2$, notice the change in the upper and lower bounds of the colour bar in Figure \ref{fig:burgers-solution-error-new} (b), which implies that methods based on autoencoders fail to accurately generate high-dimensional approximate solutions as $t$ evolves. In contrast, P-TLDINets continue to demonstrate reconstruction capabilities similar to that on the smaller datasets $\mathcal{D}_1$. 

\begin{figure}[htbp]
    \centering
	\includegraphics[width=0.7\linewidth]{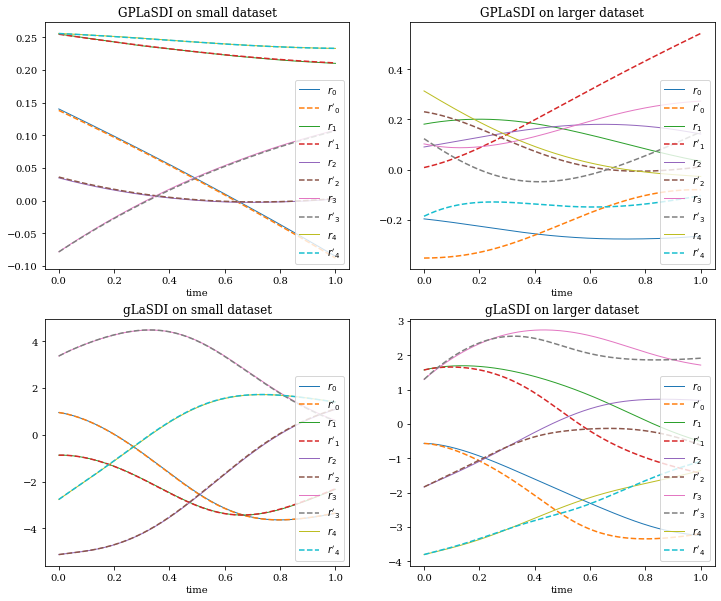}
    \caption{\textbf{2D-Burgers}: Latent states trained by GPLaSDI and gLaSDI on dataset $\mathcal{D}_1$ and $\mathcal{D}_2$. }
    \label{fig:burgers-latent-glasdi-gplasdi}
\end{figure}

In order to delve deeper into the causes of this phenomenon, Figure \ref{fig:burgers-latent-glasdi-gplasdi} depicts the low-dimensional states of the other two methods. It reveals that, in the larger parameter space $\mathcal{D}_2$, the states learned by autoencoders show more complex forms, which limit the identification ability of models, potentially causing divergence when solving the latent governing equations. Comparing Figure \ref{fig:burgers-latent-small} with Figure \ref{fig:burgers-latent-new}, it is clear that the latent performances of P-TLDINets are always smoother than that learned by autoencoders, regardless of the dimensionality of latent space. This is due to the properties of TLDNets. Therefore, this method is more generalisable.



\subsubsection{Trained on multi-scale grids}
This section demonstrates another capability of P-TLDINets, which is learning on multi-scale problems. P-PDE problems with multiple scales are commonly found in physical simulations and engineering applications\cite{wu2024multiscale, ahmed2023multiscale}. 

\begin{figure}[!t]
\centering
\begin{tabular}{ccc}
\begin{minipage}{0.3\linewidth}
\includegraphics[width = \linewidth,angle=0,clip=true]{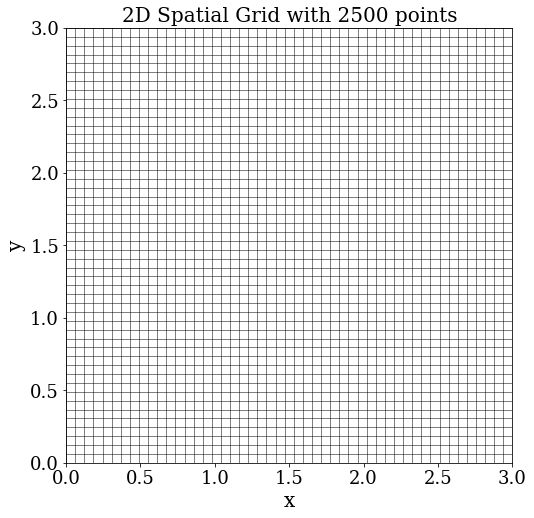}  
\end{minipage}
&
\begin{minipage}{0.3\linewidth}
\includegraphics[width = \linewidth,angle=0,clip=true]{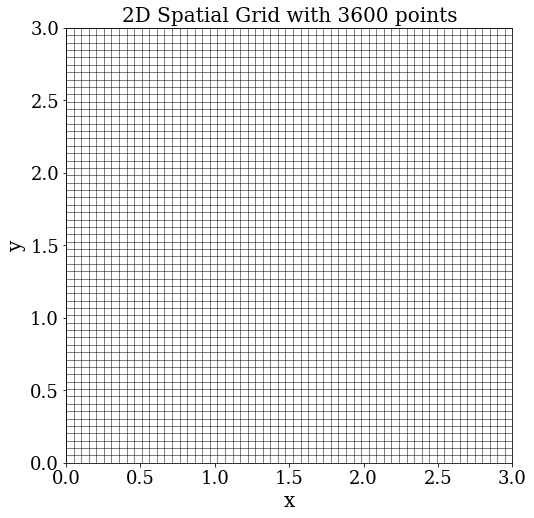}  
\end{minipage} 
&
\begin{minipage}{0.3\linewidth}
\includegraphics[width = \linewidth,angle=0,clip=true]{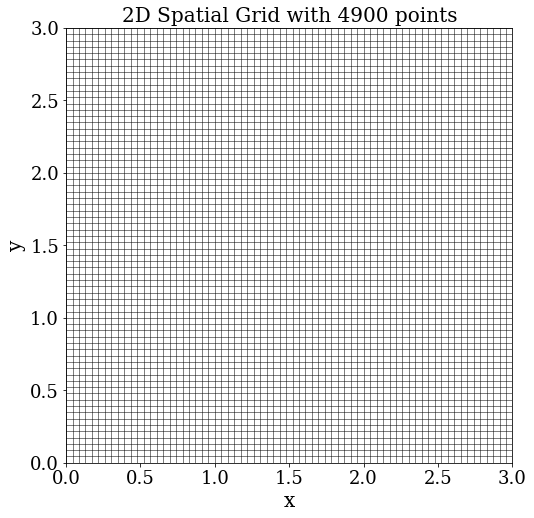}  
\end{minipage} 

 \\
(a) {\small 2500 nodes}&
(b) {\small 3600 nodes}&
(c) {\small 4900 nodes}\end{tabular}
\caption{\textbf{2D-burgers}: Three types of square grids with different scales.}
\label{fig:burgers-multiscale-grid}
\end{figure}

\begin{figure}[!ht]
    \centering
	\includegraphics[width=0.8\linewidth]{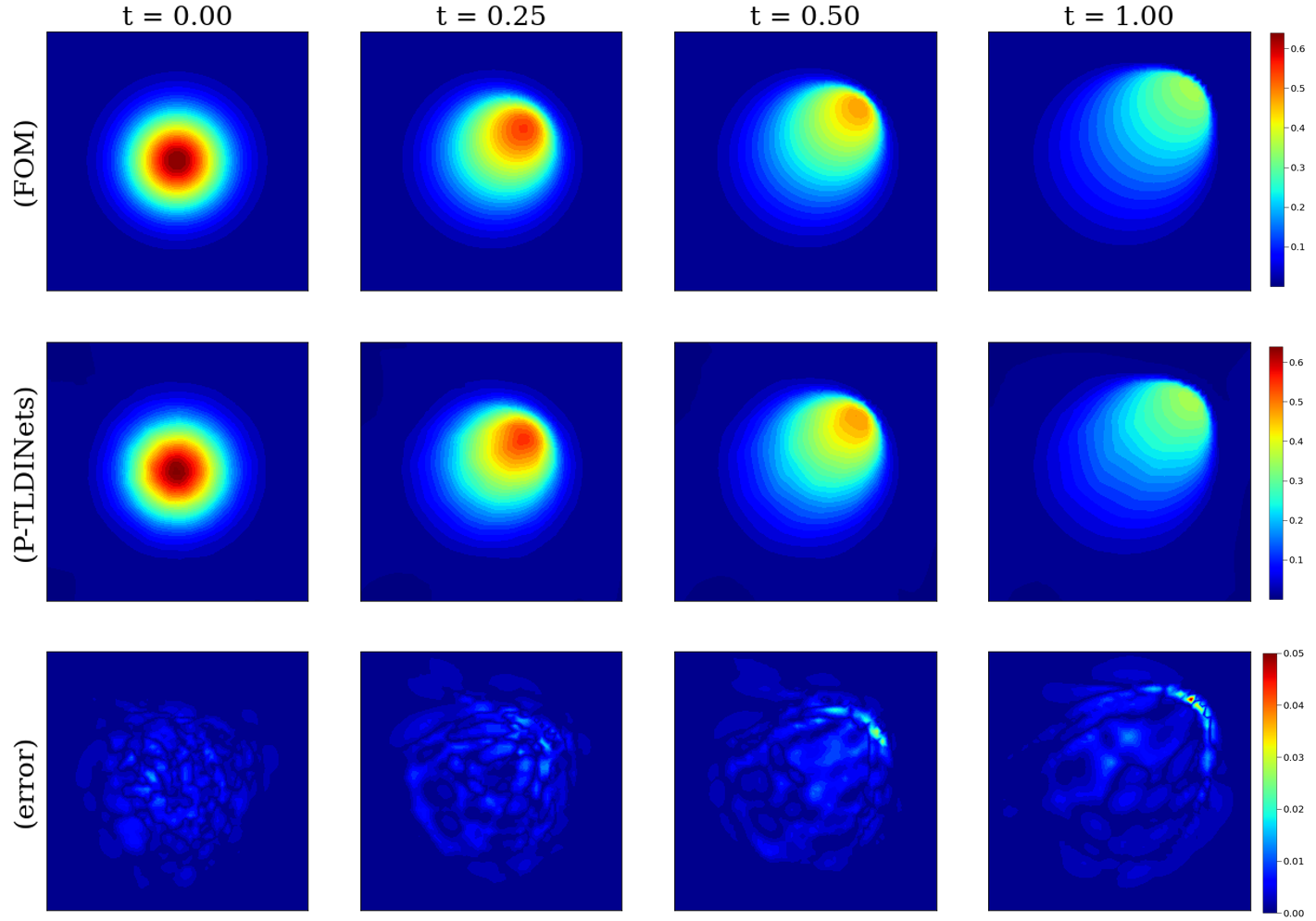}
    \caption{\textbf{2D-Burgers}: The high-fidelity velocity solutions, predicted solutions of P-TLDINets and errors for multi-scale grids at $\bm{\mu}^*=[0.64, 1.21]$. 
    } 
    \label{fig:burgers-multi}
\end{figure}

In this example, three types of square grids with different scales are fed into the model. The grids are shown in Figure \ref{fig:burgers-multiscale-grid}. 
For simplicity, these grids are uniform and divided into 50, 60 and 70 segments along $x_1$ axis and $x_2$ axis respectively, resulting in 2500, 3600, and 4900 mesh points in total. The training data $\{\bm{u}\}$ are generated according to the random selection in $\mathcal{D}_2$. The other conditions and network structure are the same as those in the previous training. The trained model is then applied to make predictions on the grid (c) in Figure \ref{fig:burgers-multiscale-grid}, which has 4900 nodes, achieving a reconstruction accuracy of 98.73\%. 
Figure \ref{fig:burgers-multi} show the specific prediction results of P-TLDINets at $\bm{\mu}^*=[0.64, 1.21]$.

\subsection{Case 2: Lock exchange}\label{sec:lock-exchange}
Lock-exchange is a classic numerical experiment in fluid mechanics, designed to analyse the mixing and flow behaviours between fluids of varying densities.
A 2D rectangular flat-bottomed tank is filled with source fluid on one side and ambient fluid on the other, separated by a vertical gate in the middle. The source fluid has lower temperature and higher density, while the ambient fluid is the opposite. After the gate is removed, the denser fluid collapses under the lighter one. The two fluids gradually mix under gravity until viscosity dominates. The density interface between the source fluid and the ambient fluid forms Kelvin–Helmholtz billows\cite{de1996Kelvin-Helmholtz}, due to instability.

Assuming the temperature on the right side is $T_0$ and on the left side is $-T_0$, the involved equations include several formulas from fluid mechanics. The incompressible fluid model can be written as:

\begin{equation}
\left\{
    \begin{aligned}
        \rho \left(\frac{\partial \textbf{u}}{\partial t} + (\textbf{u} \cdot \nabla) \textbf{u}\right) &= -\nabla p + \nu \nabla^2 \textbf{u} + \rho \textbf{g} \\[0.1cm]
        \nabla \cdot \textbf{u} &= 0 \\[0.1cm]
        \frac{\partial \bm{T}}{\partial t} + \textbf{u} \cdot \nabla \bm{T} &= \alpha \nabla^2 \bm{T} \quad (x_1, x_2, t) \in \Omega \times [0 \rm{s}, 8 \rm{s}]
    \end{aligned}
\right.
\end{equation}
where the first equation is Navier-Stokes equation, the second represents the incompressibility condition, and the third is the energy equation, also known as the heat conduction equation. $\textbf{u}$ is the velocity vector and $p$ is the pressure. $\rho$ stands for the fluid density.  $\bm{T}$ is temperature, which is the scalar data observed and learned in this case.  $\Omega = [0\,\rm{m}, 0.1\,\rm{m}] \times [0\,\rm{m}, 0.8\,\rm{m}]$. Here, thermal diffusion is not considered. 

The initial condition for velocity $\textbf{u}$ and temperature $\bm{T}$ is summarised as follows:
\begin{equation}
\left\{
    \begin{aligned}
        \textbf{u}(x_1, x_2, 0) &= 0 \\
        \bm{T}(x_1, x_2, 0) &= 
        \begin{cases} 
            -T_0, & \quad x_1 < 0.4 \, \rm{m} \\
            T_0. & \quad x_1 \geq 0.4 \, \rm{m}
        \end{cases}
    \end{aligned}
\right.
\end{equation}
The lock exchange adopts a no-slip boundary condition, meaning that the velocity remains zero at the bottom of the domain. The other relevant physical configurations are shown in Table \ref{table:le-configuration}. 

The high-fidelity solutions are obtained using an unstructured mesh finite element model, Fluidity\cite{pain2005three, Fluidity}.  
The time interval is set to $\Delta t=0.005\rm{s}$. The grid is non-uniform and holds constant over time. Considering that the flow characteristics of the lock exchange may present more dynamic variability in the middle section, a grid refinement is employed between $0.2\rm{m}\leq x \leq 0.6 \rm{m}$.

\begin{table}\label{table:le-configuration}
\centering
\begin{tabular}{ccc}
    \toprule
    Physical parameters (unit) &  Notation  &  Value \\
    \midrule
    kinematic viscosity coefficient ($\rm{m}^2\rm{s}^{-1}$)  &  $\nu$ &  $10^{-6}$ \\
    gravitational acceleration ($\rm{ms}^{-2}$)  &  $\rm{g}$  &  10 \\
    thermal diffusivity ($\rm{ms}^{-2}$) &  $\alpha$  & 0 \\
   \bottomrule
\end{tabular}
\caption{\textbf{Lock-exchange}: Configurations of simulation in Fluidity.}
\end{table}
In this case, $\bm{\mu} = \{T_0, \nu\} \in \mathcal{D} = [4.9, 5.1] \times [10^{-6}, 10^{-5}]$ forms the parameter space. 

\subsubsection{Trained on a single-scale grid}

\begin{figure}[!t]
\centering
\begin{tabular}{c}
\begin{minipage}{0.96\linewidth}
\includegraphics[width = \linewidth,angle=0,clip=true]{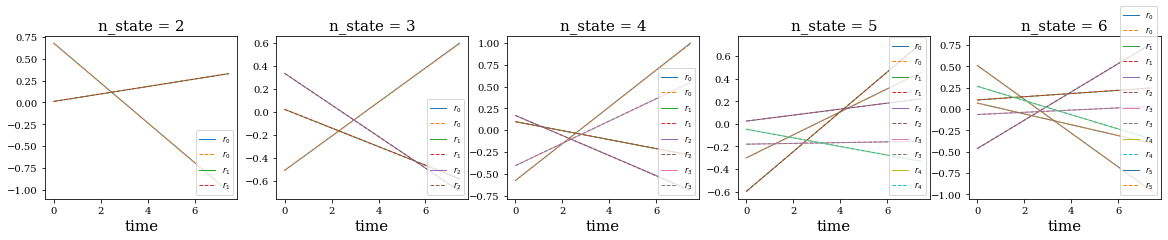}  
\end{minipage}
 \\
(a) {\small }\\
\begin{minipage}{0.96\linewidth}
\includegraphics[width = \linewidth,angle=0,clip=true]{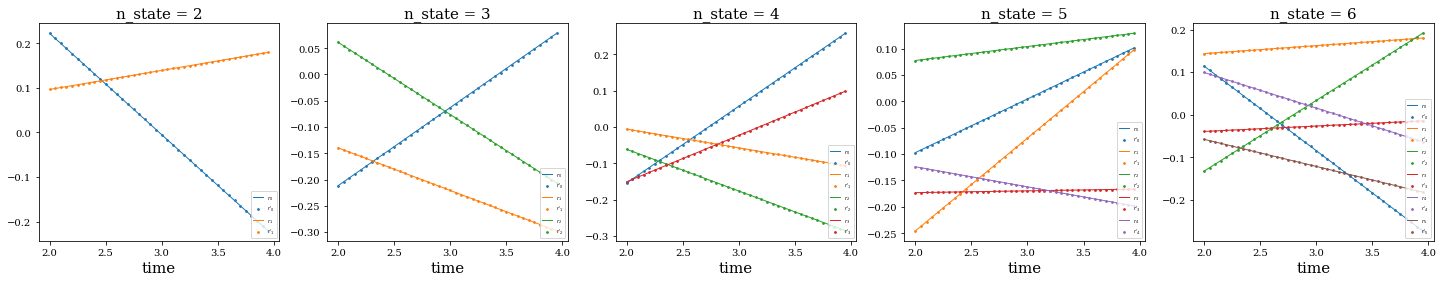}  
\end{minipage}
\\
(b) {\small}
\\
\end{tabular}
\caption{\textbf{Lock-exchange}: Low-dimensional dynamics and the solutions of the identified equations at $\bm{\mu} = [5.0, 10^{-6}]$. (a) and (b) show the dimension of latent space $N_s$ ranging from $2$ to $6$. (a) represents results for the entire time frame $t \in [0, 0.8]$, while (b) shows only in $t \in [0.2, 0.4]$. }
\label{fig:le-latent}
\end{figure} 

In this case, P-TLDINets are first trained on a single-scale grid with 1491 node points. 
25 samples are randomly selected as the training parameter set $\mathcal{D}_{train}$ and the discrete test parameter set $\mathcal{D}^h$ is made up of 225 samples. 
We trained P-TLDINets from $N_s=2$ to $N_s=6$ and the quadratic term of the time derivative in Equation \ref{equ:dynamic_state} is not involved. 
The structure of dynamical network $NN_{dyn}$ is
\begin{equation}
    \begin{aligned}
        NN_{dyn}: & \rm{Linear}(N_s+2,9) 
    \stackrel{\rm{Tanh}}{\longrightarrow} \rm{ResNetBlock}(9, 9) 
    \stackrel{\rm{Tanh}}{\longrightarrow} \rm{Linear}(9,9) 
    \stackrel{\rm{Tanh}}{\longrightarrow} \\
    & \rm{ResNetBlock}(9, 9)
    \stackrel{\rm{Tanh}}{\longrightarrow} \rm{Linear}(9,N_s).
    \end{aligned}
\end{equation}
The structure of the reconstruction network $NN_{rec}$ is 
\begin{equation}
    \begin{aligned}
        NN_{rec}: & \rm{Linear}(N_s+4,11) 
    \stackrel{\rm{Tanh}}{\longrightarrow} \rm{ResNetBlock}(11, 20) 
    \stackrel{\rm{Tanh}}{\longrightarrow} \rm{Linear}(20,20)
    \stackrel{\rm{Tanh}}{\longrightarrow}   \\
    & \rm{ResNetBlock}(20, 20)
    \stackrel{\rm{Tanh}}{\longrightarrow} \rm{Linear}(20,20)
    \stackrel{\rm{Tanh}}{\longrightarrow} \rm{ResNetBlock}(20, 11)
    \stackrel{\rm{Tanh}}{\longrightarrow}  \\
    & \rm{Linear}(20, 20)
    \stackrel{\rm{Tanh}}{\longrightarrow} \rm{ResNetBlock}(20,20)
    \stackrel{\rm{Tanh}}{\longrightarrow} \rm{Linear}(20, 20)
    \stackrel{\rm{Tanh}}{\longrightarrow}  \\
    & \rm{ResNet}(20,20)
    \stackrel{\rm{Tanh}}{\longrightarrow} \rm{Linear}(20,11)
    \stackrel{\rm{Tanh}}{\longrightarrow} \rm{Linear}(11,2).
    \end{aligned}
\end{equation}
The initial mapping network $NN_{z_0}$ has a form of:  
\begin{equation}
    \begin{aligned}
        NN_{z_0}: \rm{Linear}(2,9) 
    \stackrel{\rm{Tanh}}{\longrightarrow} \rm{Linear}(9, 9) 
    \stackrel{\rm{Tanh}}{\longrightarrow} \rm{Linear}(9, 9)
    \stackrel{\rm{Tanh}}{\longrightarrow} \rm{Linear}(9,N_s).
    \end{aligned}
\end{equation}
In the library of basis functions $\Theta(\bm{z})$, only linear and constant terms are taken into account. The optimiser is configured to Adam optimiser with a learning rate of 0.05 initially, decaying by a factor of 0.6 every 500 iterations and $N_{tol}=500$. The loss hyperparameters are set as $\omega_{DI}=0.5$, $\omega_{coef}=0$ and $\omega_{z_0}=0.5$. 

\begin{table}
\centering
\begin{tabular}{c|ccccc}
   \toprule
   $N_s$ & 2 & 3 & 4 & 5 & 6  \\
   \midrule
   $L_2$ error ($\%$)  & 0.8289  &  0.6977 & 1.310  & 0.8827  &  0.8368  \\ 
   \bottomrule
\end{tabular}
\caption{\textbf{Lock-exchange}: Total $L_2$ errors on testing samples $\mathcal{D}^h \subset \mathcal{D}_{trian}$ of different $N_s$.}
\label{table:le-error-dimenion}
\end{table}

\begin{figure}[!t]
    \centering
	\includegraphics[width=0.95\linewidth]{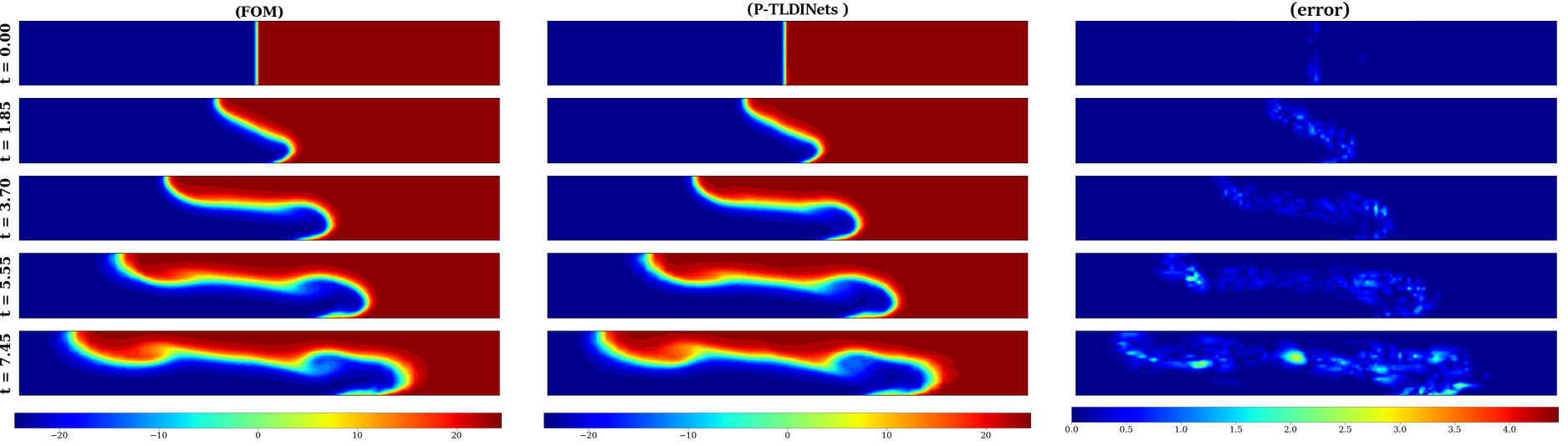}
    \caption{\textbf{Lock-exchange}: High-fidelity solutions , reconstruction results and errors of P-TLDINets at $\bm{\mu}^*=[4.929, 8.07^{-6}]$ vary with $t$.}
    \label{fig:le-solution-error}
\end{figure}

\begin{figure}[!t]
    \centering
	\includegraphics[width=0.9\linewidth]{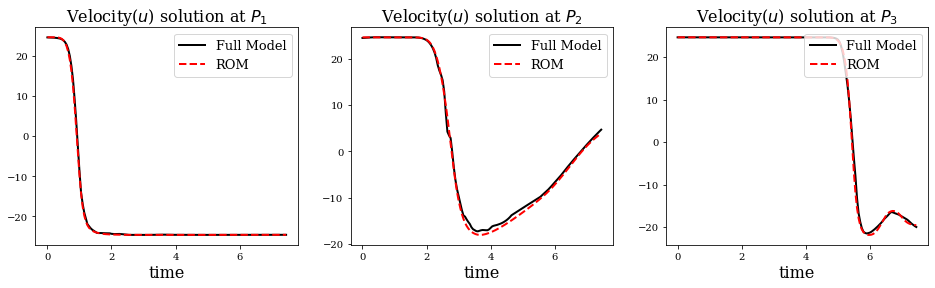}
    \caption{\textbf{Lock-exchange}: The flow fields reconstructed by P-TLDINets at $\bm{\mu}^*=[4.929, 8.07^{-6}]$ for $P_1$, $P_2$ and $P_3$.}
    \label{fig:le-p1p2p3}
\end{figure}

Figure \ref{fig:le-latent} and Table \ref{table:le-error-dimenion} illustrate the dimensionality reduction performances across various low-dimensional spaces. It is clear that P-TLDINets still show greater smoothness, even under more intricate dynamical systems in this case. This feature enhances the effectiveness of subsequent equation identification and facilitates predictions with a $176.5 \times$ speed-up.

Figure \ref{fig:le-solution-error} shows the prediction results for $\bm{\mu}^*=[4.929, 8.07^{-6}]$ in $D^h$. 
More detailed results are shown in Figure \ref{fig:le-p1p2p3}, showing the flow field reconstructions at three single points, $P_1$, $P_2$ and $P_3$, with the same $\bm{\mu}^*$. It is apparent that P-TLDINets exhibit excellent reconstruction capability.

\subsubsection{Trained on multi-scale grids}  

\begin{figure}[!t]
\centering
\begin{tabular}{c}
\begin{minipage}{0.9\linewidth}
\includegraphics[width = \linewidth,angle=0,clip=true]{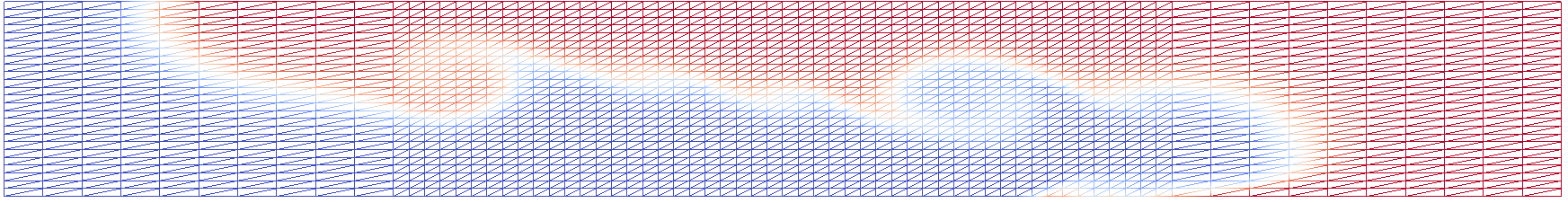}  
\end{minipage}
 \\
(a) {\small 1864 nodes}
\\
\begin{minipage}{0.9\linewidth}
\includegraphics[width = \linewidth,angle=0,clip=true]{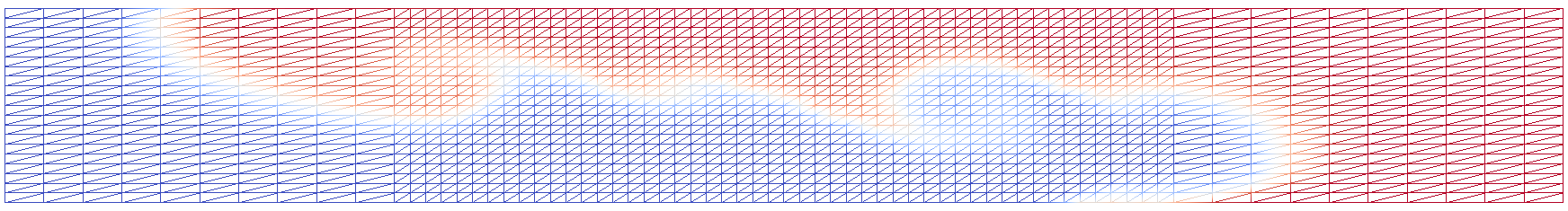}  
\end{minipage}
\\
(b) {\small 1491 nodes}
\\
\begin{minipage}{0.9\linewidth}
\includegraphics[width = \linewidth,angle=0,clip=true]{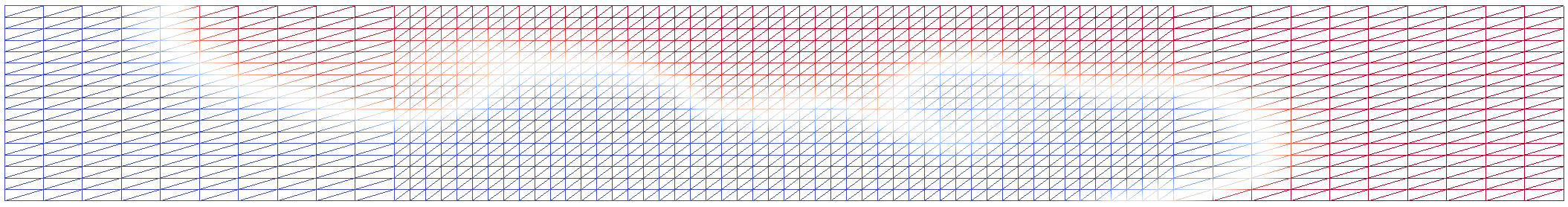} 
\end{minipage} 
\\
(c) {\small 1278 nodes}
\\
\end{tabular}
\caption{\textbf{Lock-exchange}: Three types of unstructured grids with different scales.}
\label{fig:le-multiscale-grid}
\end{figure}

\begin{figure}[!ht]
    \centering
	\includegraphics[width=1.0\linewidth]{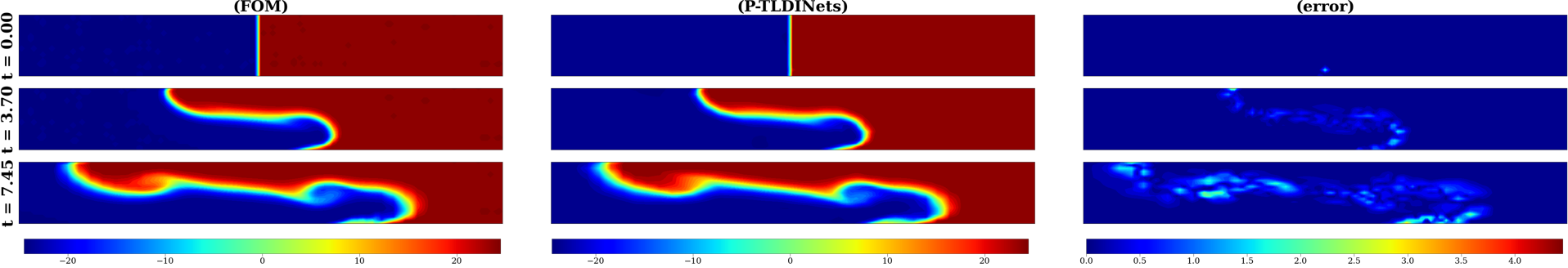}
    \caption{\textbf{Lock-exchange}: The high-fidelity temperature solutions, predicted solutions of P-TLDINets and errors for multi-scale grids at $\bm{\mu}^*=[5.00, 4.214\times 10^{-6}]$. 
    } 
    \label{fig:le-multi}
\end{figure}

In the lock-exchange scenario, grids of different scales are also tested. These grids obey the sparsity characteristics described previously and are depicted in Figure \ref{fig:le-multiscale-grid}. These three grids have 1894, 1491, and 1278 spatial points respectively. $D_{train}$ is randomly divided into three parts and each part corresponds to a different grid size. 
The high-dimensional data $\{\bm{u}\}$ are generated by Fluidity on these three grids for network training as input. Finally, predictions are made on grid (c) in Figure \ref{fig:le-multiscale-grid}. The prediction results achieves 97.00\% recovery accuracy. Figure \ref{fig:le-multi} presents the specific prediction results of P-TLDINets at $\bm{\mu}^*=[5.00, 4.214\times 10^{-6}]$. 

\section{Conclusion}\label{sec:conclusion}
\vspace{-2pt}


This paper introduces Multiscale Parametric Taylor series-based Latent Dynamics Identification Networks (P-TLDINets), a novel dimensionality reduction neural network for the purpose of robust, accurate, effective and resource-efficient data-driven computation. 
The proposed model integrates with Taylor series-based Latent Dynamic Networks Latent Dynamics Networks (TLDNets) and Identification of Dynamics (ID models), which are the core strengths in our work as well. Three sub-networks and coefficients of ID models are trained simultaneously and interactively, facilitating a simple and smoother discovery of latent dynamics. In order to enhance the exploration in parameter spaces, a state-of-the-art $k$ nearest neighbour (KNN) method based on inverse distance weighting (IDW) is used to predict approximate scalar fields of unknown parameter samples incorporating with spatial information. 

To demonstrate the utility of P-TLDINets, it has been applied to dynamical problems with nonlinear characteristics, including 2D Burgers' equations and the lock-exchange case. Numerical results show that, compared to autoencoders based latent identification methods, P-TLDINets are equipped with a lighter structure and superior performances, tending to reflect better learning capabilities in more complex parameter spaces with less training time. In addition, it is able to deal with multiscale meshes. The $L_2$ errors of the P-TLDINets remain consistently below $2\%$, achieving a significant computational speedup. 

P-TLDINets provide a transformation from leading-edge modelling reductions (such as methods based on POD and autoencoders). In detail, instead of explicitly constructing an autoencoder based on high-dimensional discrete coordinates, P-TLDINets automatically generate a latent representation of the system's states and implement a direct query for individual spatial coordinates. Additionally, it addresses the limitations of LDNets in handling problems with varying initial conditions, thereby broadening its generalisation performance. The ID model improves interpretability in low-dimensional spaces, and the concept of locally identifiable predictions makes P-TLDINets more competitive. 

The presented parametric framework is generalisable, allowing for the integration of other sampling (such as greedy sampling) and other interpolation methods. P-TLDINets, as a multiscale model actually, not only accommodate input with varying grids during training procedure, but also enhance spatial flexibility in the online reconstruction process. This enables P-TLDINets to handle multiscale meshes relatively easily, paving the way for future applications in adaptive meshing and other aspects.

\section*{Acknowledgments}
The authors acknowledge the support of the Fundamental Research Funds for the Central Universities, the Top Discipline Plan of Shanghai Universities-Class I and Shanghai Municipal Science and Technology Major Project (No. 2021SHZDZX0100), National Key $R\&D$ Program of China(NO.2022YFE0208000).	
	

	\clearpage
	\bibliographystyle{unsrt} 
	\bibliography{references}
	
\end{document}